\documentclass[10pt,twocolumn,letterpaper]{article}

\usepackage[pagenumbers]{cvpr} 

\definecolor{cvprblue}{rgb}{0.21,0.49,0.74}
\usepackage[pagebackref,breaklinks,colorlinks,allcolors=cvprblue]{hyperref}
\usepackage{eccvabbrv}

\usepackage{graphicx}
\usepackage{booktabs}
\usepackage[accsupp]{axessibility}
\usepackage{multirow}
\usepackage{array}

\usepackage{algorithm}
\usepackage{algorithmic}
\usepackage{amsmath}
\usepackage{amssymb}

\usepackage{makecell}

\usepackage[table]{xcolor}

\usepackage{orcidlink}
\usepackage{pifont}
\usepackage{fontawesome}

\usepackage{twemojis}

\usepackage{tcolorbox}
\tcbuselibrary{breakable}
\usepackage{enumitem}
\usepackage{tikz}
\usetikzlibrary{arrows.meta,positioning,shapes.geometric}

\tcbset{
    promptbox/.style={
        colback=cyan!5!white,
        colframe=cyan!50!black,
        fonttitle=\bfseries\large,
        breakable
    }
}

\renewcommand{\arraystretch}{2.0}

\usepackage{caption}

\makeatletter
\let\old@maketitle\@maketitle
\renewcommand{\@maketitle}{%
  \old@maketitle
  \vspace{-0.9em}
  \begin{center}
    \includegraphics[width=0.95\linewidth]{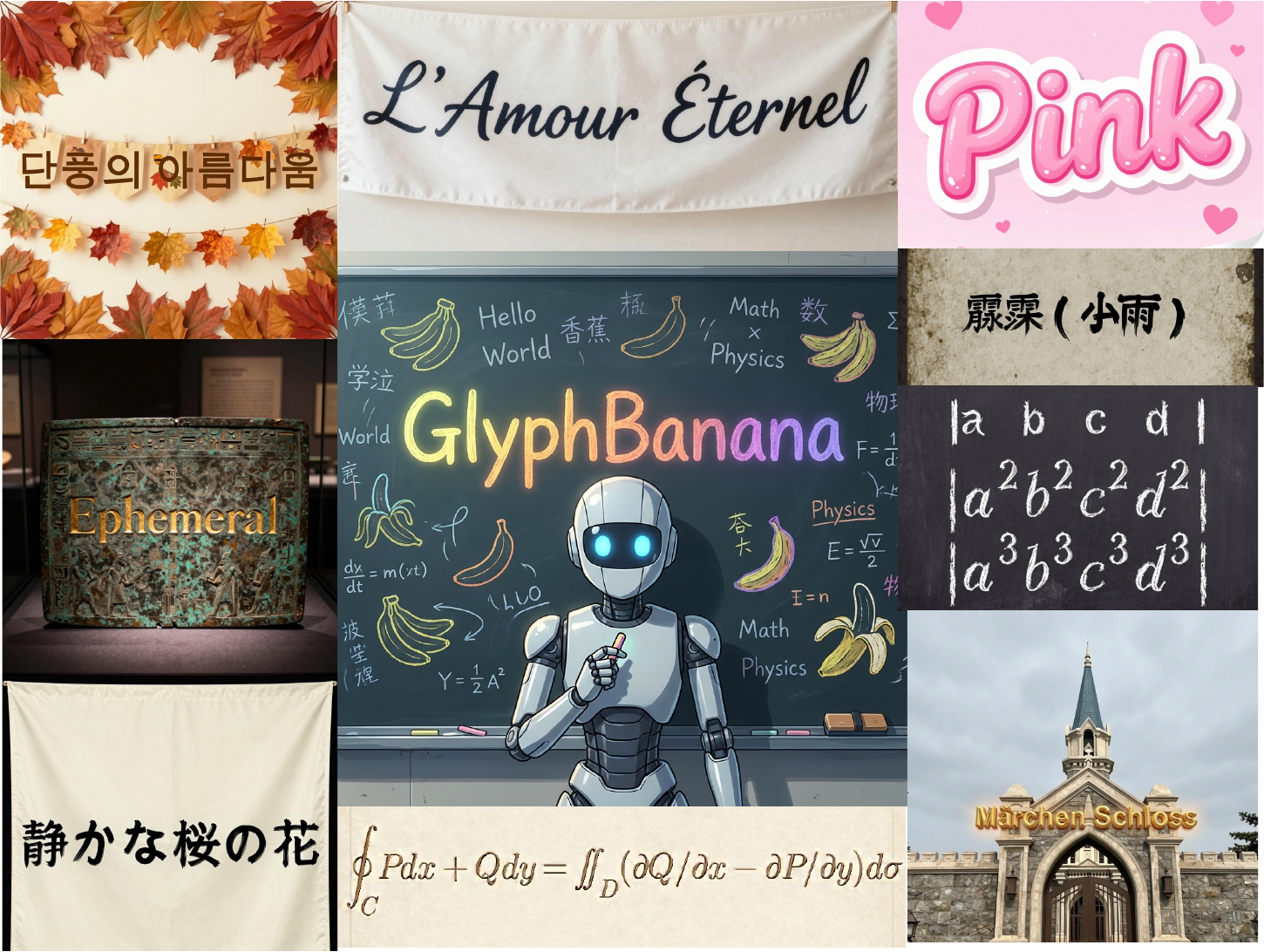}
    \captionsetup{hypcap=false}
    \captionof{figure}{Gallery of various text rendering results sampled by GlyphBanana.}
    \label{fig:teaser}
  \end{center}
  \vspace{-0.8em}
}
\makeatother

\title{\vspace{-1.5em}GlyphBanana: Advancing Precise Text Rendering Through Agentic Workflows\vspace{-1.5em}}

\author{
Zexuan Yan$^{1,2*}$ \quad
Jiarui Jin$^{2*}$ \quad
Yue Ma$^{3}$ \quad
Shijian Wang$^{2,4}$ \quad \\[-1.5em]
Jiahui Hu$^{5}$ \quad
Wenxiang Jiao$^{2}$ \quad
Yuan Lu$^{2\dagger}$ \quad
Linfeng Zhang$^{1\dagger}$\\[-1.5em]
{\small $^{1}$Shanghai Jiao Tong University \quad
$^{2}$Xiaohongshu Inc. \quad
$^{3}$Hong Kong University of Science and Technology} \\[-1.5em]
{\small $^{4}$Southeast University \quad
$^{5}$South China University of Technology}
\vspace{-1.5em}
}

\begin{document}
\maketitle
{\let\thefootnote\relax\footnotetext{$^{*}$\,Equal contribution.}\footnotetext{$^{\dagger}$\,Corresponding author.}}
\vspace{-1.5em}

\begin{abstract}
Despite recent advances in generative models driving significant progress in text rendering, accurately generating complex text and mathematical formulas remains a formidable challenge. This difficulty primarily stems from the limited instruction-following capabilities of current models when encountering out-of-distribution prompts. To address this, we introduce GlyphBanana, alongside a corresponding benchmark specifically designed for rendering complex characters and formulas. GlyphBanana employs an agentic workflow that integrates auxiliary tools to inject glyph templates into both the latent space and attention maps, facilitating the iterative refinement of generated images. Notably, our training-free approach can be seamlessly applied to various Text-to-Image (T2I) models, achieving superior precision compared to existing baselines. Extensive experiments demonstrate the effectiveness of our proposed workflow. Associated code is publicly available at \url{https://github.com/yuriYanZeXuan/GlyphBanana}.
\end{abstract}

\begin{figure*}[htbp]
    \centering
    \includegraphics[width=0.9\linewidth]{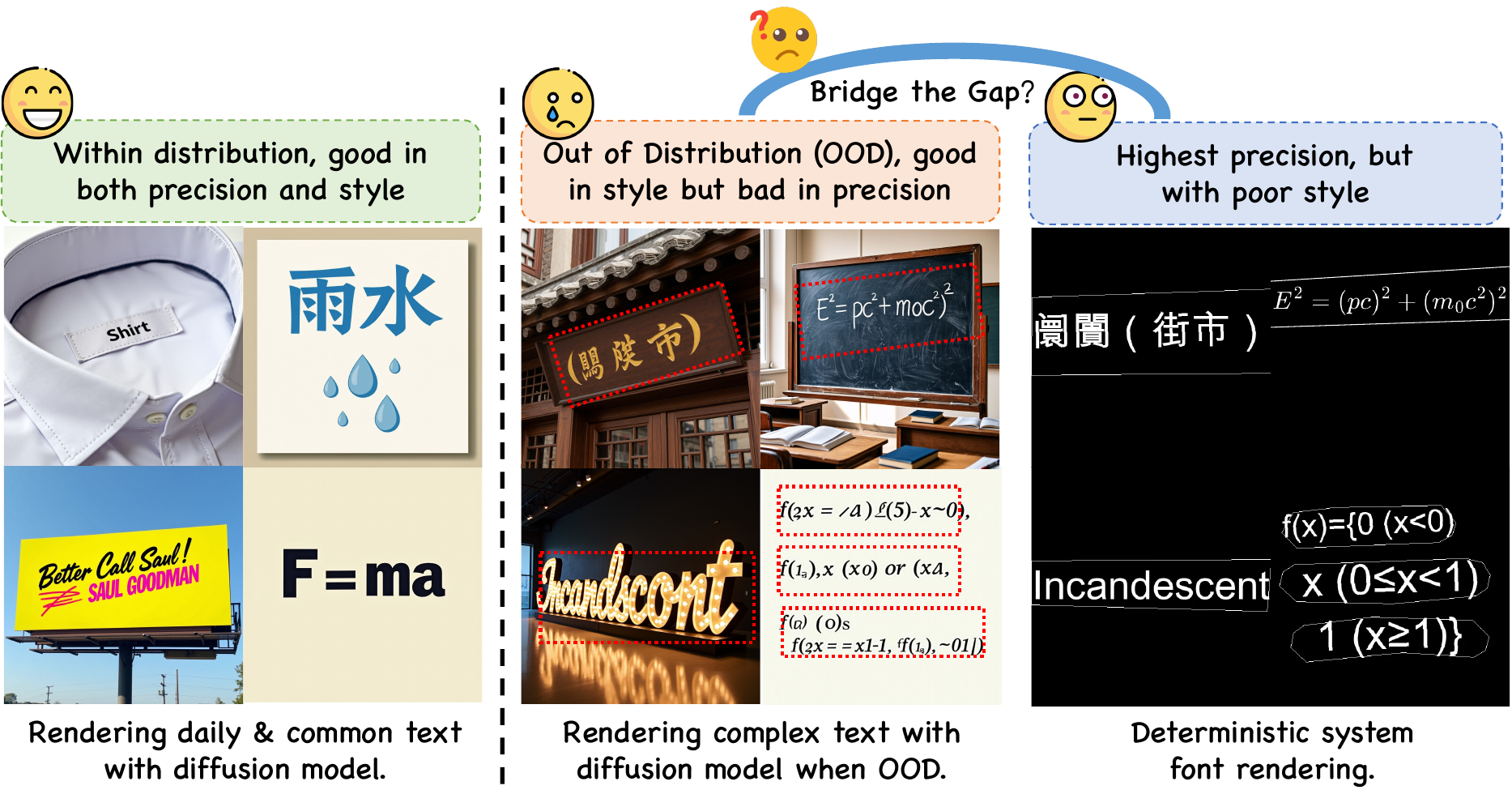}
    \caption{\textbf{The illustration of motivation.} We observe that while in-distribution cases show satisfying precision-style banlance, there exists huge gap between OOD cases and deterministic rendered texts.}
    \label{fig:motivation}
\end{figure*}

\section{Introduction}
\label{sec:intro}

Recent diffusion transformers~\cite{peebles2023scalablediffusionmodelstransformers,  ma2024followpose, ma2025followcreation, ma2026fastvmt, ma2025followyourmotion, ma2025controllable, esser2024scalingrectifiedflowtransformers} have demonstrated remarkable progress in image generation, driving a wide range of applications such as commercial advertising, poster design, and scientific visualization. 
In these contexts, accurate text rendering plays a critical role, imposing stringent demands on both the generalizability of diffusion models and their capacity for multilingual instruction following. 
Basic mainstreaming generative models, such as Z-Image \cite{imageteam2025zimageefficientimagegeneration} and Qwen-Image \cite{wu2025qwenimagetechnicalreport}, excel at rendering frequently encountered text, including short English phrases, common everyday Chinese expressions, and simple mathematical equations. But they perform poorly on rare English words, complex Chinese characters, and sophisticated scientific formulas (as exemplified in Figure~\ref{fig:motivation}).

To improve their precise text rendering performance, existing approaches can be broadly categorized into two paradigms, namely training-based and training-free methods.
Training-based approaches, such as GlyphByT5~\cite{liu2024glyph} and FluxText~\cite{lan2025flux}, adopt strategies of either LoRA-based fine-tuning or fine-tuning on the text encoder. 
Despite their effectiveness in certain scenarios, these methods commonly suffer from limited generalization ability and a heavy reliance on high-quality annotated datasets.
Training-free methods, such as TextCrafter~\cite{textcrafter} and FreeText \cite{freetext}, typically incorporate a glyph prior as a spatial layout constraint to regulate and guide text rendering. 
However, an overly strong glyph prior tends to disrupt the background and overall visual style of the image, resulting in style inconsistency between the rendered text and its surrounding content.
We also note that system font tools offer high-precision text rendering capabilities, yet lack flexibility, as they require handcrafted designs to adapt to specific styles.

\begin{figure*}[t]
    \centering
    \includegraphics[width=1\linewidth]{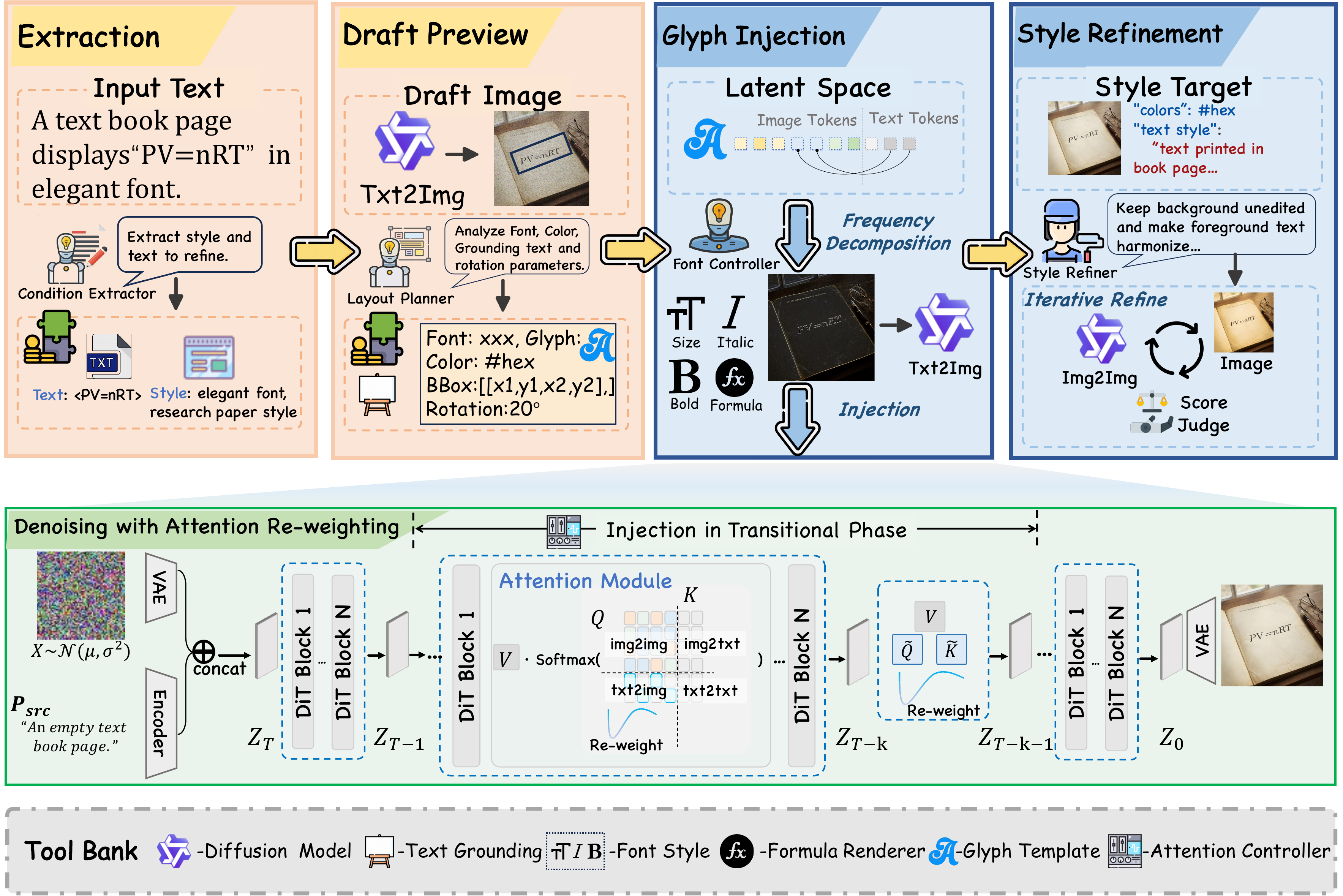}
    \caption{\textbf{Overview of the GlyphBanana agentic pipeline.} The workflow comprises four stages: (1)~\textit{Extraction Stage}  parses the input into text content and style attributes; (2)~\textit{Draft Preview Stage} generates an initial image via a Layout Planner; (3)~\textit{Glyph Injection Stage}  applies Frequency Decomposition in latent space and Attention Re-weighting inside each DiT block; (4)~\textit{Style Refinement Stage} employs iterative refinement with a Style Refiner and Score Judger. The bottom panel details the denoising process with the Attention Re-weighting.}
    \label{fig:pipeline}
\end{figure*}

In this paper, we propose a novel agentic workflow, termed \textbf{GlyphBanana}, which effectively integrates the precise rendering capabilities of system font rendering tools with the generative flexibility of diffusion models, thereby enabling autonomous adaptation to arbitrary styles without requiring any manual design intervention.
Specifically, GlyphBanana operates through the following four sequential stages. In the \textbf{extraction stage}, GlyphBanana first employs vision-language models to extract the target text content and the desired rendering style from the input prompt. 
Subsequently, in the \textbf{draft preview stage}, text-to-image models are applied to generate a preliminary image in the desired style as a reference preview, which is followed by a layout planner equipped with text grounding tools to produce a glyph template that encapsulates detailed attributes, including font type, color, bounding box coordinates, and rotation parameters.  
The \textbf{glyph injection stage} constitutes the core component of GlyphBanana, wherein the produced glyph template is integrated into the generative model through both latent space and attention modules. 
Specifically, for the latent space, frequency decomposition is employed to disentangle the denoising representations of the glyph template into low- and high-frequency components, after which the information-dense high-frequency components are injected into the latent space. 
For the attention modules, an attention re-weighting mechanism is introduced to incorporate the glyph template as a bias term into the attention maps within each DiT block. 
Finally, in the \textbf{style refinement stage}, the intermediate image generated from the glyph injection stage are iteratively refined by jointly optimizing the refinement prompts and the generated images to further enhance overall image quality.
It is worth noting that GlyphBanana is a training-free framework orchestrated by a collection of plug-and-play tools, enabling seamless integration with arbitrary generative models.

Existing text-rendering benchmarks~\cite{anytext,textcrafter,glyphdraw,liu2023character,Xomni}, are narrowly focused on common English words or Chinese characters, systematically neglecting rare characters and complex scientific formulas. To address this limitation, we introduce \textbf{GlyphBanana-Bench}, a comprehensive text-rendering benchmark that, to the best of our knowledge, is the first to systematically evaluate text rendering across a diverse spectrum of difficulty levels and linguistic domains, ranging from simple common words and rare Chinese characters to complex multiline scientific formulas. GlyphBanana-Bench is constructed through a combination of community-forum crawling and synthesis via Kimi-K2.5~\cite{kimik2.5}, ensuring both diversity and scalability of the benchmark data. Extensive experiments demonstrate that our GlyphBanana achieves substantial improvements in OCR accuracy, attaining scores of \textbf{85.9} (\textbf{+19.6\%}) on Z-Image and \textbf{75.8} (\textbf{+6.91\%}) on Qwen-Image, while simultaneously enhancing precision and style.



\section{Related work}
\label{sec:relatedwork}

\noindent \textbf{DiT for Image Generation\& Editing.}
Diffusion Transformer (DiT)~\cite{peebles2023scalablediffusionmodelstransformers} has emerged as an alternative to U-Net~\cite{ronneberger2015unetconvolutionalnetworksbiomedical} for image generation and editing. 
Building upon this, recent works~\cite{esser2024scalingrectifiedflowtransformers,  ma2025followfaster, ma2024followyouremoji, ma2025followyourclick, shen2025follow, chen2025contextflow, wang2024taming, feng2025dit4edit,wang2024cove, zhu2024instantswap, 
imageteam2025zimageefficientimagegeneration,
eedit, z1, z2, z3, wang2025multishotmaster, 
wu2025qwenimagetechnicalreport,
cao2025hunyuanimage, lu2025easytext, shi2024fonts, shi2025wordcon, 
labs2025flux1kontextflowmatching,
liu2025step1xeditpracticalframeworkgeneral,
Xomni} integrate Flow Matching~\cite{lipman2023flowmatchinggenerativemodeling} to improve training stability and inference efficiency.
Benefiting from its unified attention architecture, DiT also demonstrates strong capabilities in image editing. Existing approaches can be broadly categorized into single-turn and multi-turn paradigms. 
Single-turn methods, such as GLIDE\cite{nichol2022glidephotorealisticimagegeneration}, MagicBrush\cite{Zhang2023MagicBrush}, Prompt-to-Prompt\cite{hertz2022prompt}, UltraEdit\cite{gu2025ultraedittrainingsubjectmemoryfree}, and FireEdit\cite{zhou2025fireeditfinegrainedinstructionbasedimage}, perform instruction-guided edits in a one-shot manner. 
In contrast, multi-turn systems\cite{Gemini25FlashImage_Google_2025, GPTImage1_OpenAI_2025} enable iterative, context-aware editing through interactive feedback.

\noindent \textbf{Visual Text Rendering.}
Although diffusion-based methods can generate high-quality images, rendering text in images remains a challenging problem due to the need for accurate spelling, layout coherence, and style consistency. One line of work\cite{Balaji2022eDiffITD, liu2023character, Saharia2022PhotorealisticTD} leverages large language models\cite{Raffel2019ExploringTL,xue2022byt5tokenfreefuturepretrained} to improve spelling accuracy in generative models. 
Another line\cite{glyphdraw, Chen2023TextDiffuserDM, yang2024glyphcontrol} focuses on explicitly controlling text layout and content during generation.
Recent works further improve rendering quality from multiple perspectives. 
TextCenGen\cite{liang2025textcengenattentionguidedtextcentricbackground} and TextCrafter\cite{textcrafter} enhance layout and attribute consistency, while Calligrapher\cite{ma2025calligrapherfreestyletextimage} and TextMaster\cite{yan2025textmasterunifiedframeworkrealistic} focus on style control via glyph- and feature-level guidance. 
SceneVTG\cite{Zhu2024VisualTG} adopts a planning–rendering pipeline with Vision Language Models to ensure semantically coherent text.

\noindent \textbf{Image Rendering with Agentic Workflow.}
Beyond single-step generation, real-world design tasks~\cite{jiang2025personalized} often require multi-step reasoning, iterative refinement, and human-like decision making. PosterGen\cite{zhang2025postergen} simulates a design team with specialized agents for layout and styling, Agent Banana\cite{ye2026agentbananahighfidelityimage} proposes a hierarchical planner-executor framework with long-horizon memory and layer-wise manipulation.  
For image and video restoration, MoA-VR and AgenticIR\cite{liu2025moavrmixtureofagentsallinonevideo, zhu2025intelligentagenticcompleximage} extend agentic workflows to VLM-integrated multi-agent repair frameworks. In more complex settings such as creative photo retouching and task-oriented restoration, systems like JarvisIR, JarvisArt, 4KAgent, and JarvisEvo\cite{jarvisir2025, jarvisart2025,
JarvisEvo_arXiv_2025, zuo20254kagent} further demonstrate the effectiveness of agentic pipelines. 
Complementary to these system-level designs, EditThinker \cite{li2025editthinkerunlockingiterativereasoning} focuses on enhancing intra-agent capability by formulating image editing as an explicit iterative reasoning process.
\section{Preliminaries}
\label{sec:prelim}

\subsection{Multimodal Diffusion Transformer}
The Multimodal Diffusion Transformer (MM-DiT) models the generation of an image $I$ conditioned on a text prompt $P$ within a latent space. First, a pre-trained Variational Autoencoder (VAE) compresses the image into a low-dimensional latent representation $z_0 = \text{VAE}_{\text{enc}}(I)$. 

Following standard diffusion models, a forward process gradually corrupts the data $z_0$ into Gaussian noise by adding noise $\epsilon \sim \mathcal{N}(0, \mathbf{I})$. The noisy latent $z_t$ at timestep $t$ is defined as:
\begin{equation}
    z_t = \alpha_t z_0 + \sigma_t \epsilon,
\end{equation}
where $\alpha_t$ and $\sigma_t$ are the noise schedule parameters. The training objective $\mathcal{J}$ is to learn a neural network, parameterized by $\theta$, to reverse this process by predicting the added noise:
\begin{equation}
    \min \mathcal{J}_\theta =    \min \mathbb{E}_{t, z_0, \epsilon} \left[ \left\| \epsilon_\theta(z_t, t, P) - \epsilon \right\|_2^2 \right].
\end{equation}

To parameterize the denoiser $\epsilon_\theta(z_t, t, P)$ using the MM-DiT architecture, the continuous latent states and discrete text condition must be transformed into sequence representations. The noisy latent $z_t$ is spatially patchified and linearly projected to form the image tokens $\mathbf{X}_{img} = \text{Patchify}(z_t)$. Simultaneously, the text prompt $P$ is mapped by a pre-trained text encoder into a sequence of text tokens $\mathbf{X}_{txt} = \text{TextEnc}(P)$. The two modality-specific token sequences are then concatenated along the sequence dimension to construct the joint hidden state for the Transformer network:
\begin{equation}
    \mathbf{H} = [\,\mathbf{X}_{img} \,\|\, \mathbf{X}_{txt}\,].
\end{equation}
After passing through the stacked MM-DiT blocks, the updated visual components of $\mathbf{H}$ are separated and un-patchified back to the original spatial shape to yield the final noise prediction $\epsilon_\theta$.

\section{Methods}
\label{sec:methods}

\begingroup
\renewcommand{\algorithmicdo}{}
\renewcommand{\algorithmicthen}{}
\newcommand{\righttricomment}[1]{\item[]\hfill$\triangleright$~#1}
\begin{algorithm}[htbp]
    \caption{Injection with Attention Enhancement}
    \label{alg:main}
    \small
    \begin{algorithmic}[1]
    \REQUIRE Typography plan $\mathcal{P}$, Prompt $T$, Total steps $N$, Injection window $[\tau_{start}, \tau_{end})$, Bias scales $(0<s^{-}<1<s^{+})$.
    \ENSURE Glyph-injected latent $z_{0}$.
    
    \STATE $I \leftarrow \text{FontRender} (\mathcal{P})$; \quad $M \leftarrow \text{Otsu}(I)$
    \righttricomment{\textbf{Stage 1:} \textit{Preprocessing}}
    \STATE $\mathcal{I}_{txt} \leftarrow \text{FindTokenIndices}(T, \text{quoted})$
    \STATE $\mathcal{I}_{img} \leftarrow \{i \mid M[i] > 0\}$; \quad $\tilde{\mathcal{I}}_{img} \leftarrow \{i \mid M[i] = 0\}$
    \righttricomment{\textit{glyph / non-glyph indices}}
    \STATE $\tilde{z}_{list}: \{\tilde{z_0}, \ldots, \tilde{z_N}\} \leftarrow \text{Inversion}(\text{VAE}(I))$
    \righttricomment{\textit{fusion glyph template list}}
    
    \FOR{denoising step $t = N, \ldots, 1$ \textbf{do}}
        \FOR{$AttnProcessor_i$ $\in$ DiT Block \textbf{do}}
            \righttricomment{\textbf{Stage 2:} \textit{Attn. re-weighting}}
            \STATE $B \leftarrow \mathbf{0}$; \quad $\alpha^{+} \leftarrow \log(s^{+})$; \quad $\alpha^{-} \leftarrow \log(s^{-})$
            \STATE $B[\mathcal{I}_{img}, \mathcal{I}_{txt}] \mathrel{+}= \alpha^{+}$; \; $B[\mathcal{I}_{txt}, \mathcal{I}_{img}] \mathrel{+}= \alpha^{+}$
            \righttricomment{\textit{enhance}}
            \STATE $B[\tilde{\mathcal{I}}_{img}, \mathcal{I}_{txt}] \mathrel{+}= \alpha^{-}$; \; $B[\mathcal{I}_{txt}, \tilde{\mathcal{I}}_{img}] \mathrel{+}= \alpha^{-}$
            \righttricomment{\textit{suppress}}
            \STATE $Q,K,V \leftarrow \text{Linear}(h_{i,t})$; \quad $\hat{Q},\hat{K} \leftarrow \text{RoPE}(Q,K)$
            \STATE $h_{i,t-1} \leftarrow \text{SDPAttention}(\hat{Q},\hat{K},V,\; \text{bias}\!=\!B)$
            \righttricomment{\textit{SDPAttention definition Eq.~\eqref{eq:sdp_attention}}}
        \ENDFOR
        \IF{$t/N \in [\tau_{start}, \tau_{end})$ \textbf{then}}
            \righttricomment{\textbf{Stage 3:} \textit{Latent Injection}}
            \STATE $\tilde{z_{tpl}} \leftarrow \tilde{z}_{list}[t]$
            \STATE $z_{t} \leftarrow \mathcal{F}_{\text{F.D.}}(z_{t+1},\; \tilde{z_{tpl}},\; M)$
            \righttricomment{\textit{Frequency Decomposition Eq.~\eqref{eq:fd}}}
        \ENDIF
        \STATE $z_{t-1} \leftarrow \text{Scheduler.step}(z_{t}, t)$
    \ENDFOR
    
    \RETURN $z_{0}$
    \end{algorithmic}
    \end{algorithm}
\endgroup
    
\begin{figure*}[htbp]
    \centering
    \includegraphics[width=1\linewidth]{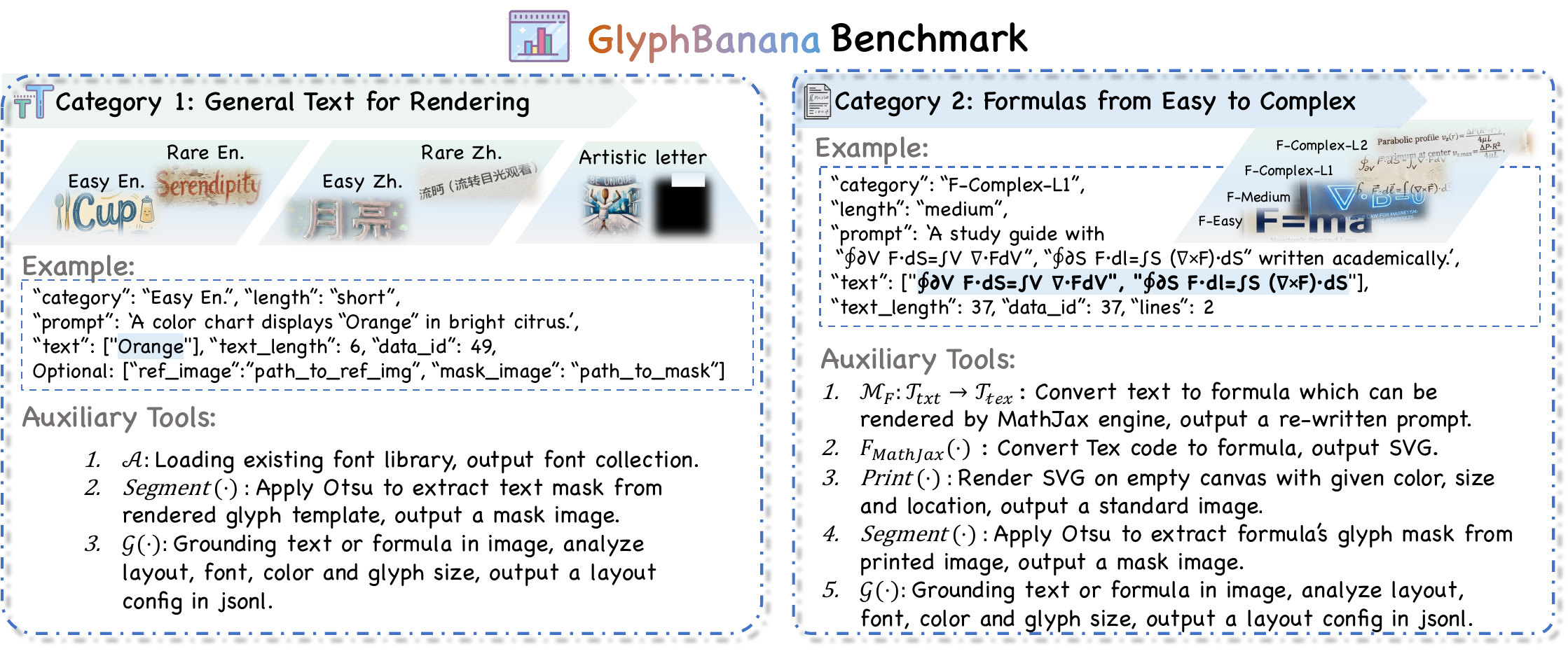}
    \caption{\textbf{Illustration of the GlyphBanana-Benchmark with auxiliary tools.} The proposed benchmark consists of two categories. General Text for Rendering assesses standard and stylized text rendering. Formulas from Easy to Complex evaluates formula rendering across varying complexities}
    \label{fig:gb-bench}
\end{figure*}
\begin{table*}[htbp]
    \caption{\textbf{Comparison of different text-rendering datasets.}  Num. refers to the number of samples in the dataset, and Avg.L refers to the average length of the char to be rendered in the dataset. {\includegraphics[height=0.3cm]{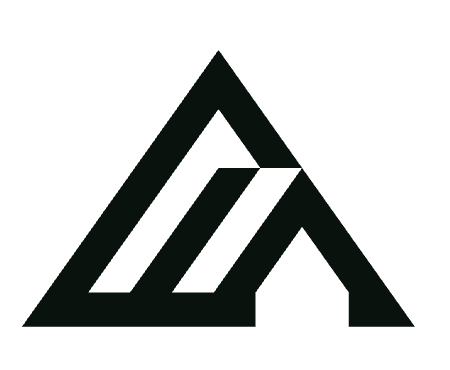}} refers to FLUX.2-klein-9B and {\includegraphics[height=0.3cm]{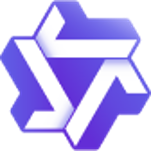}} refers to Qwen-Image-2512.}
    \centering
    \renewcommand{\arraystretch}{1.1}
    \resizebox{0.8\textwidth}{!}{
        \begin{tabular}{m{0.24\linewidth} m{0.05\linewidth} m{0.05\linewidth} m{0.06\linewidth} m{0.08\linewidth} m{0.08\linewidth} m{0.07\linewidth} m{0.07\linewidth} m{0.06\linewidth} m{0.06\linewidth} m{0.06\linewidth} m{0.06\linewidth}}
        \toprule
        \multicolumn{1}{c}{\multirow{2}{*}{\textbf{Datasets}}} & 
        \multicolumn{3}{c}{\textbf{Text Type}} & 
        \multicolumn{2}{c}{\textbf{Condition}} & 
        \multicolumn{2}{c}{\textbf{Statistics}} & 
        \multicolumn{2}{c}{\textbf{OCR Score}} &
        \multicolumn{2}{c}{\textbf{Style Score}} \\
        
        \cmidrule(lr){2-4} \cmidrule(lr){5-6} \cmidrule(lr){7-8} \cmidrule(lr){9-10} \cmidrule(lr){11-12}
        
        \multicolumn{1}{c}{} & 
        \multicolumn{1}{c}{\textbf{En.}} 
        & \multicolumn{1}{c}{\textbf{Zh.}} 
        & \multicolumn{1}{c}{\textbf{Formulas}} 
        & \multicolumn{1}{c}{\textbf{Image}} 
        & \multicolumn{1}{c}{\textbf{Mask}} 
        & \multicolumn{1}{c}{\textbf{Num.}} 
        & \multicolumn{1}{c}{\textbf{Avg.L}} 
        & \multicolumn{1}{c}{\includegraphics[height=0.3cm]{Fig/bfl.png}} 
        & \multicolumn{1}{c}{\includegraphics[height=0.3cm]{Fig/qwen.png}} 
        & \multicolumn{1}{c}{\includegraphics[height=0.3cm]{Fig/bfl.png}} 
        & \multicolumn{1}{c}{\includegraphics[height=0.3cm]{Fig/qwen.png}} \\
        \midrule
        
        DrawTextExt \cite{liu2023character} & 
        \multicolumn{1}{c}{\ding{51}} & \multicolumn{1}{c}{\ding{51}} & 
        \multicolumn{1}{c}{\ding{55}} & 
        \multicolumn{1}{c}{\ding{55}} & 
        \multicolumn{1}{c}{\ding{55}} & 
        \multicolumn{1}{c}{$220$} & \multicolumn{1}{c}{$17.0$} & 
        \multicolumn{1}{c}{0.76} & \multicolumn{1}{c}{0.81} &
        \multicolumn{1}{c}{0.83} & \multicolumn{1}{c}{0.80} \\
        
        AnyText \cite{anytext} & 
        \multicolumn{1}{c}{\ding{51}} & \multicolumn{1}{c}{\ding{51}} & 
        \multicolumn{1}{c}{\ding{55}} & 
        \multicolumn{1}{c}{\ding{55}} & 
        \multicolumn{1}{c}{\ding{55}} & 
        \multicolumn{1}{c}{$1000$} & \multicolumn{1}{c}{$21.8$} & 
        \multicolumn{1}{c}{0.33} & \multicolumn{1}{c}{0.44} &
        \multicolumn{1}{c}{0.67} & \multicolumn{1}{c}{0.69} \\
        
        CVTG-2K \cite{textcrafter} & 
        \multicolumn{1}{c}{\ding{51}} & \multicolumn{1}{c}{\ding{55}} &  \multicolumn{1}{c}{\ding{55}} & 
        \multicolumn{1}{c}{\ding{55}} & \multicolumn{1}{c}{\ding{55}} & 
        \multicolumn{1}{c}{$2000$} & \multicolumn{1}{c}{$39.5$} & 
        \multicolumn{1}{c}{0.49} & \multicolumn{1}{c}{0.51} &
        \multicolumn{1}{c}{0.75} & \multicolumn{1}{c}{0.67} \\
        
        LongText-Bench \cite{Xomni} & 
        \multicolumn{1}{c}{\ding{51}} & \multicolumn{1}{c}{\ding{51}} & 
        \multicolumn{1}{c}{\ding{55}} & 
        \multicolumn{1}{c}{\ding{55}} & 
        \multicolumn{1}{c}{\ding{55}} & 
        \multicolumn{1}{c}{$320$} & \multicolumn{1}{c}{$116.7$} & 
        \multicolumn{1}{c}{0.38} & \multicolumn{1}{c}{0.72} &
        \multicolumn{1}{c}{0.69} & \multicolumn{1}{c}{0.74} \\
        
        Ours & 
        \multicolumn{1}{c}{\ding{51}} & \multicolumn{1}{c}{\ding{51}} & 
        \multicolumn{1}{c}{\ding{51}} & 
        \multicolumn{1}{c}{\ding{51}} & 
        \multicolumn{1}{c}{\ding{51}} & 
        \multicolumn{1}{c}{$290$} & \multicolumn{1}{c}{$32.7$} & 
        \multicolumn{1}{c}{$0.37$} & \multicolumn{1}{c}{$0.71$} &
        \multicolumn{1}{c}{$0.68$} & \multicolumn{1}{c}{$0.71$} \\
        
        \bottomrule
        
        \end{tabular}
    }
    \label{table:datasets_compare}
    \end{table*}

As illustrated in Fig.~\ref{fig:pipeline}, our agentic workflow comprises four tightly coordinated stages: (1)~Extraction, which parses the user input into text content and style attributes; (2)~Draft Preview, which generates a preliminary image and derives a typography plan; (3)~Glyph Injection, which integrates precise glyph information via Frequency Decomposition and Attention Re-weighting; and (4)~Style Refinement, which iteratively improves visual harmony. The injection procedure is formalised in Algorithm~\ref{alg:main}. We detail each stage below.

\subsection{Extraction Stage}
\label{sec:extraction}
Given the user prompt ${P}_{user}$, an  \textbf{extractor} decomposes it into two components: the target text content $T$ to be rendered, and a style description $\mathcal{S}$ that characterises the desired visual appearance.
This stage output reference ground-truth for identifying text to be rendered in subsequent stages.

\subsection{Draft Preview Stage}
\label{sec:draft}
The draft preview stage produces a preliminary image and a detailed typography plan that guides glyph injection. In this stage, a draft image $I_{draft}$ is generated according to the original prompt $P_{user}$, which is analysed by
a Layout Planner, powered by the VLM equipped with text grounding tools. The planner creates a typography plan detailing the font, color, bounding boxes, and rotation angles for the generated text. This information is forwarded to the next stage to construct the injection template.

\subsection{Glyph Injection Stage}
\textbf{Formula Renderer} produces pixel-accurate glyph images via \LaTeX{} compilation for formulas, while for regular text, a \textbf{Font Controller} selects the appropriate font family,  weight, and size according to typography plan $\mathcal{P}$ from last stage, outputing accurate glyph template image $I$. Along with rendered formulas, they are encoded by VAE into a strong glyph prior $z_{tpl}$ as a template. 

\subsubsection{Frequency Decomposition.}

Frequency Decomposition is used to strengthen the high frequency structure of the glyph template in denoising latent by precisely injecting high frequency glyph details. We define the frequency-decomposed blending function as follows:

\begin{equation}
\mathcal{F}_{\text{F.D.}}(z, z_{tpl}, M) = \text{LF}(z) + \text{HF}(z)\!\odot\!(1\!-\!M) + \text{HF}(z_{tpl})\!\odot\!M,
\label{eq:fd}
\end{equation}
where $\text{LF}(z)\!=\!\text{GaussianBlur}(z)$ extracts the low frequency component, $\text{HF}(z)\!=\!z - \text{LF}(z)$ extracts the high frequency residual, and $M$ is the mask that specifies the glyph-covered tokens.  Specifically, GaussianBlur is implemented by a Gaussian blur kernel to do average pooling in latent space on image which is rendered using system font by font controller agent according to the typography plan. For the mask $M$, we use Otsu's~\cite{otsu1979threshold} method to segment the image into foreground and background. Since directly injecting the glyph latent into the denoising latent may lead to artifacts, we set a injection window $[\tau_{start}, \tau_{end})$ to control the injection timing, leaving space for adjusting edge smoothness and style consistency with the background for diffusion model. This part is applied at each denoising step $t \in[\tau_{start}, \tau_{end})$ in the Algorithm~\ref{alg:main}, stage 3.

\subsubsection{Injection with Attention Enhancement.}
We introduce a technique called Glyph Injection to inject the glyph latent into the denoising latent as shown in the Algorithm~\ref{alg:main}, stage 2. As inspired by P2P~\cite{hertz2022prompt}, TextCrafter~\cite{textcrafter}, manipulating the attention value in the attention processors of DiT blocks could effectively affect response of output response pattern to prompt tokens. 
Specifically, we inject the Self-Attention module within the DiT block. Following the standard attention formulation in Transformers \cite{transformer}, the computation incorporating a bias matrix $B$ is expressed as follows:
\begin{equation}
    \text{SDPAttention}(\hat{Q}, \hat{K}, V, B) = \text{softmax}\left( \frac{\hat{Q}\hat{K}^\top}{\sqrt{d}} + B \right) V,
    \label{eq:sdp_attention}
\end{equation}
where $\hat{Q}$, $\hat{K}$, and $V$ denote the projected query, key, and value matrices respectively. $d$ is the scaling dimension, and $B$ is the attention bias matrix designed for explicit re-weighting. The matrix $B$ is initialized to zero, and its non-zero elements $B_{i,j}$ are assigned by glyph-template in latent space, illustrated in Fig.~\ref{fig:pipeline}, stage 3.
We use the glyph latent template $\tilde{z}_{tpl}$ to specify image tokens that are likely to be affected by the glyph injection, which can be divided into two parts: glyph-covered tokens, defined by Indices $\mathcal{I}_{img}$ and non-glyph-covered tokens, defined by Indices $\tilde{\mathcal{I}}_{img}$. Similarly, text tokens are extracted and its corresponding indices are defined as $\mathcal{I}_{txt}$. For precisely control the attention computing process, we enhance the attention map value from $\mathcal{I}_{img}$ to $\mathcal{I}_{txt}$ and suppress value from $\mathcal{I}_{txt}$ to $\tilde{\mathcal{I}}_{img}$ in the attention processors of DiT blocks.



\subsection{Style Refinement Stage}
\subsubsection{Iterative Refinement.}
To improve text rendering quality and ensure stylistic harmony with the background, we introduce an Iterative Refine module. As illustrated in Fig.~\ref{fig:pipeline}, this module utilizes a pretrained image-to-image diffusion model $\mathcal{F}_{\text{DM}}$ to refine the output of the Glyph Inject stage. The refinement is driven by a VLM that serves dual functions: a \textbf{Style Refiner} that identifies and corrects discordant visual attributes (\eg, color, texture, shadow) based on intermediate outputs and produces an amended prompt $P'$, and a \textbf{Score Judger} that evaluates each candidate and selects the optimal result.

Formally, given the injected image $I_{\text{origin}}$, its typography plan prompt $P$, and the glyph mask $M$, we construct a diverse candidate pool from three refinement strategies:
$I_{\text{mask}} = M \odot I_{\text{origin}} + (1\!-\!M) \odot \mathcal{F}_{\text{DM}}(I_{\text{origin}} \mid P)$ which restricts regeneration to the non-glyph region to preserve text contours,
$I_{\text{ref}} = \mathcal{F}_{\text{DM}}(I_{\text{origin}} \mid P,\, M)$ which conditions on $M$ as a reference to guide generation while allowing broader stylistic adjustment, 
and $I_{\text{sty}} = \mathcal{F}_{\text{DM}}(I_{\text{origin}} \mid P',\, M)$ where $P'$ is the amended prompt produced by the Style Refiner.
The Score Judger then selects the best output from the candidate pool:
\begin{equation}
I^{*} = \arg\max_{I \,\in\, \{I_{\text{origin}},\, I_{\text{mask}},\, I_{\text{ref}},\, I_{\text{sty}}\}} \; \mathcal{S}_{\text{VLM}}(I, P),
\label{eq:score_judger}
\end{equation}
where $\mathcal{S}_{\text{VLM}}$ denotes the VLM-based quality assessment. The system operates in a closed loop: the Style Refiner analyzes $I^{*}$, updates $P'$, regenerates the candidate pool, and the Score Judger re-evaluates, iterating until convergence or a maximum number of rounds is reached.

\begin{figure*}[htbp]
    \centering
    \includegraphics[width=0.9\linewidth]{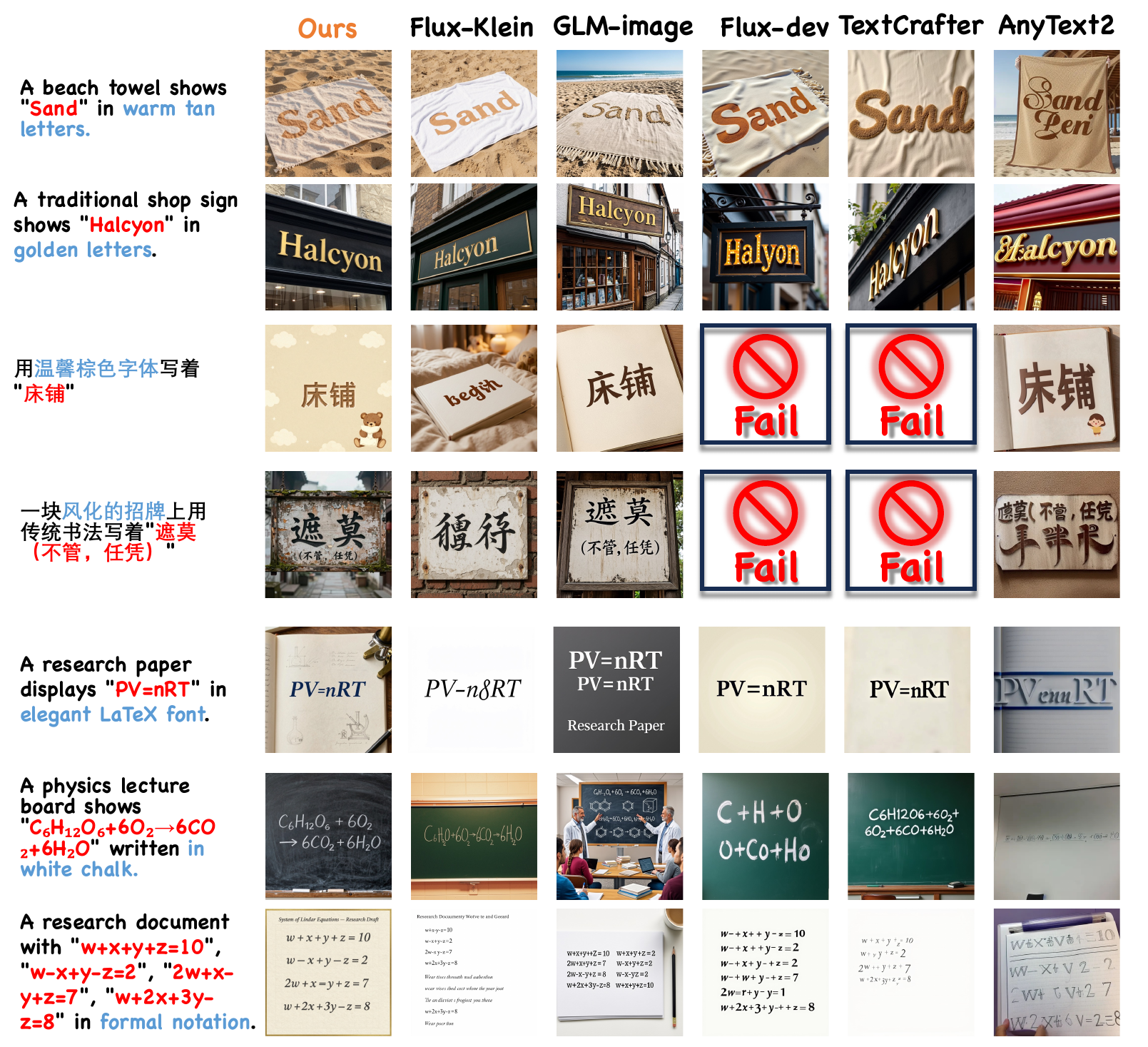}
    \caption{\textbf{Qualitative comparisons with other baselines}. \textcolor{red}{Fail} denotes the FLUX.1-dev based models unable to follow instructions to render chinese text due to its limited text-encoder. Besides, we color the quoted text in red, referring to the target text to be rendered, and color the style text related to the glyph in blue.}
    \label{fig:qualitative_comp}
\end{figure*}
\begin{table*}[h]
    \caption{Quantitative comparison results for text-rendering metrics.}
    \label{tab:quantitative_ablation}
    \centering
    \renewcommand{\arraystretch}{1.1} 
    \setlength{\tabcolsep}{5pt}       
    
\resizebox{0.9\textwidth}{!}{
\begin{tabular}{llcccccccc}
\toprule
\multicolumn{2}{c}{\multirow{2}{*}{\textbf{Method}}} & \multicolumn{2}{c}{\textbf{OCR Score$\uparrow$}} & \multicolumn{2}{c}{\textbf{VLM Score$\uparrow$}} & \multicolumn{2}{c}{\textbf{ITM Score$\uparrow$}} & \multicolumn{2}{c}{\textbf{User Study$\downarrow$}} \\
\cmidrule(lr){3-4} \cmidrule(lr){5-6} \cmidrule(lr){7-8} \cmidrule(lr){9-10} 

\multicolumn{2}{c}{} & \textbf{Acc.} & \textbf{Ned.} & \textbf{Style} & \textbf{Faith.} & \textbf{VQA} & \textbf{CLIP} & \textbf{Aesthetic} & \textbf{Faith.} \\ 
\midrule

\midrule 
\multirow{4}{*}{$Ours_{+Z-Image}$} 
                        & w/o re-weight    & 72.3 & 76.7 & 0.745 & 0.704 & 0.808 & 0.709 & 2.73 & 2.70 \\
                        & w/o refine & 84.0 & 86.7 & 0.725 & 0.745 & 0.798 & 0.710 & 3.47 & 3.40 \\
                        & w/o F.D.    & 84.5 & 87.3 & 0.755 & 0.764 & 0.803 & 0.708 & 2.73 & 2.83 \\
    \rowcolor{gray!20}  & full        & \textbf{85.9} & \textbf{88.1} & \textbf{0.765} & \textbf{0.764} & \textbf{0.814} & \textbf{0.720} & \textbf{1.07} & \textbf{1.07} \\
    
\midrule
\multirow{4}{*}{$Ours_{+QwenImage}$}
                          & w/o re-weight    & 70.7 & 74.8 & 0.676 & 0.689 & 0.814 & 0.680 & 2.43 & 2.23 \\
                          & w/o refine & 75.7 & 79.6 & 0.687 & 0.777 & 0.820 & \textbf{0.697} & 3.60 & 3.43 \\
        & w/o F.D.        & 74.5 & 78.7 & 0.718 & 0.812 & 0.819 & 0.696 & 2.93 & 3.00 \\
\rowcolor{gray!20}    & full    & \textbf{75.8} & \textbf{79.9} & \textbf{0.729} & \textbf{0.830} & \textbf{0.839} & 0.694 & \textbf{1.03} & \textbf{1.33} \\
\bottomrule
\end{tabular}
}
\end{table*}

\section{Benchmark and Evaluation Protocals}

\subsection{Benchmark}
Current evaluation frameworks for text-rendering diffusion models inadequately assess out-of-vocabulary (OOV) tokens, complex notation, and the hierarchical multiline layouts typical of scientific equations. To bridge this evaluation gap, we present the GlyphBanana-Benchmark as illustrated in Fig.~\ref{fig:gb-bench}.  We delicately collect and construct a wide range of text and formulas to be rendered along with supplementary tools for agentic workflow. For category of general text, we provide auxiliary tools including font library, segmentation and text-grounding tools for specifying text font and layout. To the best of our knowledge, it is the first benchmark to systematically evaluate text rendering capabilities across a comprehensive difficulty spectrum ranging from simple words to complex, multiline mathematical formulas, while supporting multimodal inputs and auxiliary rendering tools. The dataset is meticulously constructed: the rare Chinese word subset is curated by crawling community forums~\cite{zhihu_jingluo_2020}, whereas the English and complex formula subsets are entirely synthesized using Kimi-K2.5~\cite{kimik2.5}.

Furthermore, we conduct quantitative evaluations with other similar benchmark on text type, input conditions, statistics of benchmark size, and score related to precision and style metrics.
Specifically, we employ two popular open-source diffusion models to assess existing baseline metrics, which are {\includegraphics[height=0.3cm]{Fig/bfl.png}}FLUX.2-klein-9B and {\includegraphics[height=0.3cm]{Fig/qwen.png}}Qwen-Image-2512 in Table.~\ref{table:datasets_compare}. Results reveal that accurately rendering rare Chinese characters and complex formulas remains a challenge for current diffusion-based methods. More qualitative results refer to supplementary materials.

\begin{table*}[h]
  \caption{Quantitative comparison results for text-rendering metrics. {\includegraphics[height=0.3cm]{Fig/bfl.png}} represents for FLUX.2-klein-9B and {\includegraphics[height=0.3cm]{Fig/qwen.png}} represents for Qwen-Image-2512.}
  \label{tab:quantitative_comparison}
  \centering
  \renewcommand{\arraystretch}{1.1}
  \setlength{\tabcolsep}{5pt}
  \resizebox{0.9\textwidth}{!}{
\begin{tabular}{llcccccccc}
\toprule
\multicolumn{2}{c}{\multirow{2}{*}{\textbf{Method}}} & \multicolumn{2}{c}{\textbf{OCR Score$\uparrow$}} & \multicolumn{2}{c}{\textbf{VLM Score$\uparrow$}} & \multicolumn{2}{c}{\textbf{ITM Score$\uparrow$}} & \multicolumn{2}{c}{\textbf{User Study$\downarrow$}} \\
\cmidrule(lr){3-4} \cmidrule(lr){5-6} \cmidrule(lr){7-8} \cmidrule(lr){9-10}
\multicolumn{2}{c}{} & \textbf{Acc.} & \textbf{Ned.} & \textbf{Style} & \textbf{Faith.} & \textbf{VQA} & \textbf{CLIP} & \textbf{Aesthetic} & \textbf{Faith.} \\ 
\midrule
\multicolumn{2}{l}{AnyText2} & 33.8 & 40.5 & 0.661 & 0.438 & 0.641 & 0.637 & 7.80 & 7.62 \\
\multicolumn{2}{l}{TextCrafter} & 34.0 & 39.6 & 0.672 & 0.371 & 0.804 & 0.680 & 6.75 & 6.47 \\
\midrule
\multirow{2}{*}{Flux.1} & .dev & 27.9 & 34.3 & 0.691 & 0.280 & 0.771 & 0.639 & 6.75 & 7.03 \\
                      & FluxText & 25.0 & 28.4 & 0.600 & 0.351 & 0.718 & 0.656 & 6.98 & 6.83 \\
\multirow{1}{*}{Flux.2} & .klein\;{\includegraphics[height=0.3cm]{Fig/bfl.png}}  & 36.7 & 42.1 & 0.676 & 0.521 & 0.821 & 0.686 & 6.38 & 6.08 \\
\midrule
\multicolumn{2}{l}{GLM-Image}  & 62.1 & 70.8 & 0.728 & 0.700 & 0.807 & 0.681 & 5.50 & 5.77 \\
\multicolumn{2}{l}{Zimage} & 71.8 & 76.3 & \cellcolor{green!20}\underline{0.750} & 0.703 & 0.813 & \cellcolor{blue!20}\textbf{0.723} & 5.07 & 5.12 \\
\multicolumn{2}{l}{Qwen-Image\;{\includegraphics[height=0.3cm]{Fig/qwen.png}}} & 70.9 & 74.7 & 0.705 & \cellcolor{green!20}\underline{0.767} & \cellcolor{blue!20}\textbf{0.840} & 0.699 & 4.63 & 4.98 \\
\midrule
\multicolumn{2}{l}{$Ours_{+zimage}$} 
& \cellcolor{blue!20}\textbf{85.9} {\textcolor{green!60!black}{\tiny$\uparrow$14.1}} 
& \cellcolor{blue!20}\textbf{88.1} {\textcolor{green!60!black}{\tiny$\uparrow$11.8}} 
& \cellcolor{blue!20}\textbf{0.765} {\textcolor{green!60!black}{\tiny$\uparrow$0.015}}
& 0.764 {\textcolor{green!60!black}{\tiny$\uparrow$0.061}}
& 0.814 {\textcolor{green!60!black}{\tiny$\uparrow$0.001}}
& \cellcolor{green!20}\underline{0.720} {\textcolor{gray!60!black}{\tiny$\downarrow$0.003}}
& \cellcolor{blue!20}\textbf{2.27} & \cellcolor{green!20}\underline{2.58} \\
\multicolumn{2}{l}{$Ours_{+QwenImage}$}
& \cellcolor{green!20}\underline{75.8} {\textcolor{green!60!black}{\tiny$\uparrow$4.9}}
& \cellcolor{green!20}\underline{79.9} {\textcolor{green!60!black}{\tiny$\uparrow$5.2}}
& 0.729 {\textcolor{green!60!black}{\tiny$\uparrow$0.024}}
& \cellcolor{blue!20}\textbf{0.830} {\textcolor{green!60!black}{\tiny$\uparrow$0.063}}
& \cellcolor{green!20}\underline{0.839} {\textcolor{gray!60!black}{\tiny$\downarrow$0.001}}
& 0.694 {\textcolor{gray!60!black}{\tiny$\downarrow$0.005}}
& \cellcolor{green!20}\underline{2.87} & \cellcolor{blue!20}\textbf{2.52} \\
\bottomrule
\end{tabular}
}
\end{table*}
\subsection{Evaluation Protocols}
We adopt a multi-dimensional evaluation protocol covering Optical Character Recognition(OCR) Score,  Vision-Language Model(VLM) Score,  Image-Text Matching(ITM) Score, and User Study.
\textbf{OCR Score} represents the precision of the rendered text, here we use OCR Accuracy(OCR-Acc) and OCR Normalized Edit Distance(OCR-NED) to evaluate the precision of the rendered text. Define $d(g,p)$ as the Levenshtein distance between the ground-truth text $g$ and the predicted text $p$. OCR-Acc~$= 1 - d(g,p)/|g|$, is a recall-oriented score that quantifies how much of the ground-truth text $g$ is correctly rendered in the prediction $p$. OCR-NED~$= 1 - d(g,p)/\max(|g|,|p|)$, is a symmetric similarity that additionally penalizes hallucinated text. 
Following recent practice, we additionally query a VLM to obtain \textbf{VLM Score}, including VLM-Style related to clarity, coherence, and aesthetics and VLM-Faithfulness related to scene, object, style, and text placement adherence to the prompt.
\textbf{Image-Text Matching} measures the alignment between the rendered text and the reference image, using CLIP Score~\cite{hessel2021clipscore} and VQA Score~\cite{lin2024evaluating}.  \textbf{User Study} is evaluated by performing a human preference sheet on aesthetic and faithfulness preferences to evaluate the text rendering quality, ordering images from best to worst.

\begin{figure*}
    \centering
    \includegraphics[width=1\linewidth]{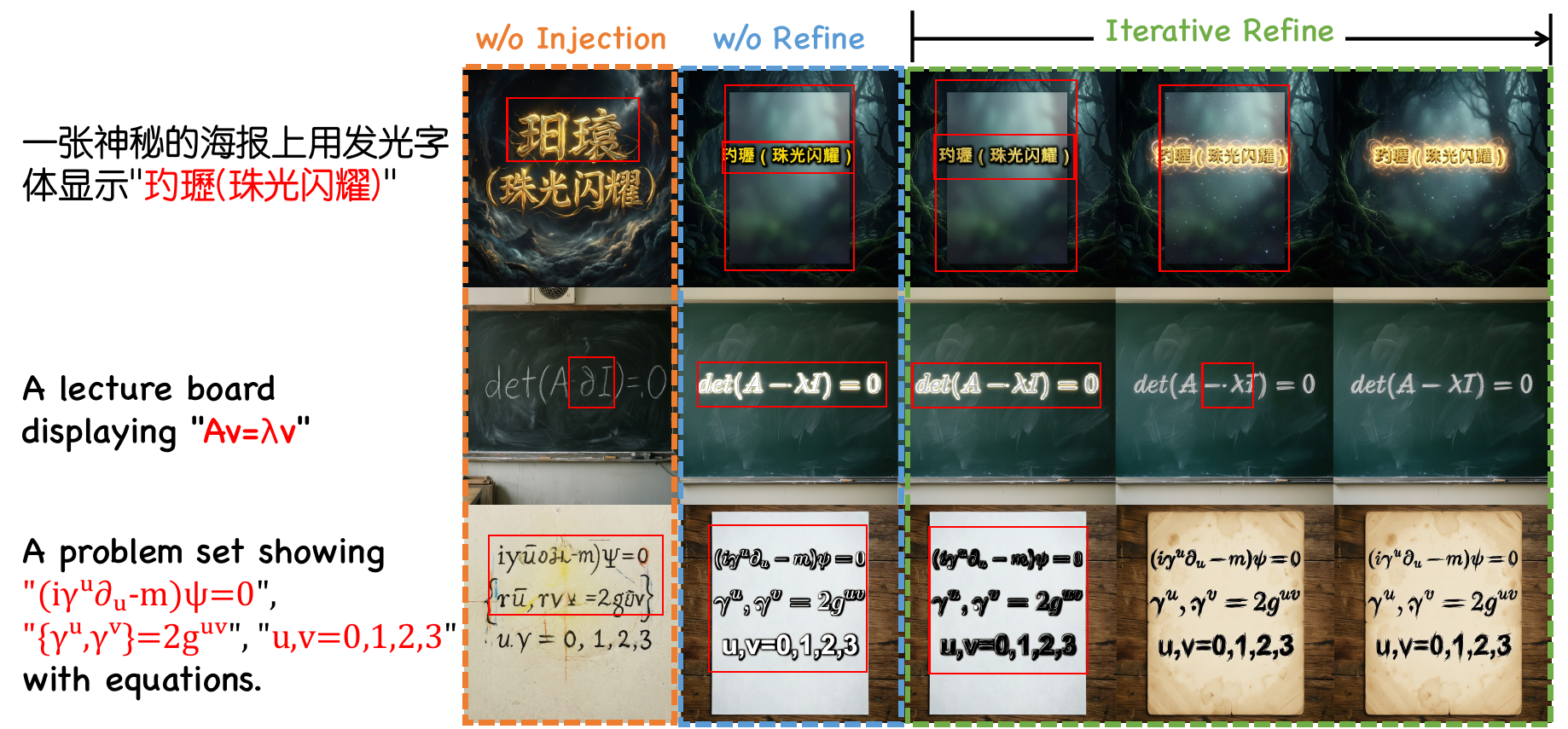}
    \caption{Qualitative comparison results.}
    \label{fig:qualitative_ablation}
\end{figure*}
\section{Experiments}
\label{sec:exp}

\subsection{Implementation Details}
In our experiments, For Text-to-Image Generation, we adopt two open-source diffusion backbones: QwenImage-2512 with 50 denoising steps and ZImage-turbo with 20 denoising steps. For Style-Refiner, FLUX.2-klein-9B model is adopted for Image-to-Image Generation. To support the agent's core planning and evaluation, we employ the powerful open-source Qwen3-VL-235B-A22B-Instruct model, which serves as the Layout Planner during the Draft Preview stage, Style Refiner and Score Judger during the Style Refine stage, OCR excecutor and VLM Score evaluator for evaluation.  All experiments are conducted on NVIDIA H800 GPUs. During Glyph Injection, attention re-weighting is applied over denoising timesteps in the bias of scaled-dot-production-attention in the range $(0.2, 0.8)$, with 2.0 enhancement scale and 0.1 for attention suppression. More details are provided in the supplementary materials.

\subsection{Comparison with baselines}

\noindent\textbf{Qualitative Comparison.} 
We conducted extensive experiments with our proposed Agentic method (Z-Image by default) on GlyphBanana-Benchmark, comparing it with other approaches.
Specifically, we comprehensively compare the text rendering capabilities of the models on simple to rare English, Chinese, and from simple to complex multiline formulas.
As shown in Figure \ref{fig:qualitative_comp}, our method achieves superior performance in text rendering on simple to rare English, Chinese, and from simple to complex multiline formulas compared to existing methods. For FLUX.1-dev and TextCrafter, Chinese rendering is not supported due to its limited text-encoder. For multiline formulas, other methods show lower precision or duplicate rendering. Our method supports multi-language, and shows highest precision for rendering formulas.

\noindent\textbf{Quantitative Comparison.}
We performed comprehensive quantitative experiments on GlyphBanana-Benchmark. Our methods significantly improve the metrics related to rendering precision and quality score, including User Study, achieving the highest text accuracy among all other methods including text-rendering specific approaches such as AnyText2 and TextCrafter. Compared to the original text rendering baseline, the T2I matching metrics are nearly identical, but the style and faithfulness scores are higher including User Study, getting the best overall performance. 

\begin{figure}
    \centering
    \includegraphics[width=1.0\linewidth]{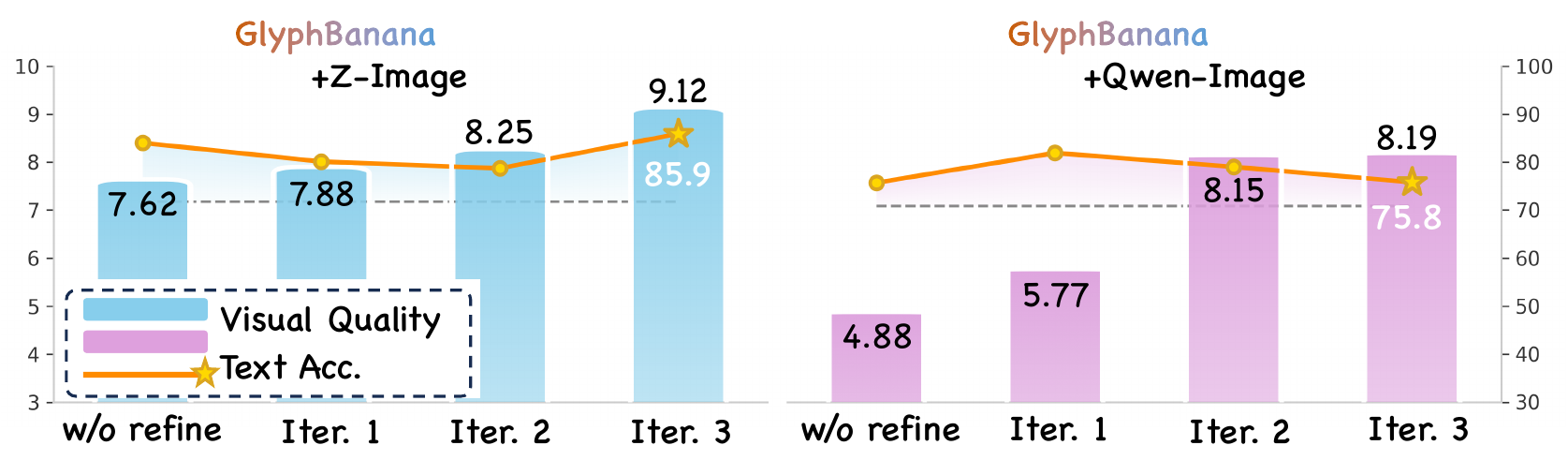}
    \caption{Metric comparisons for multi-turn refinement.}
    \label{fig:quantitative_refine}
\end{figure}

\subsection{Ablation Study}
\label{sec:ablation2}
Extensive qualitative and quantitative experiments are conducted using text accuracy, image quality, and user study metrics to validate the effectiveness of three key operations in our agentic workflow: Frequency Decomposition (F.D. for short), Attention Enhancement(re-weight for short), and Iterative Refine(refine for short).

\begin{figure}
    \centering
    \includegraphics[width=\linewidth]{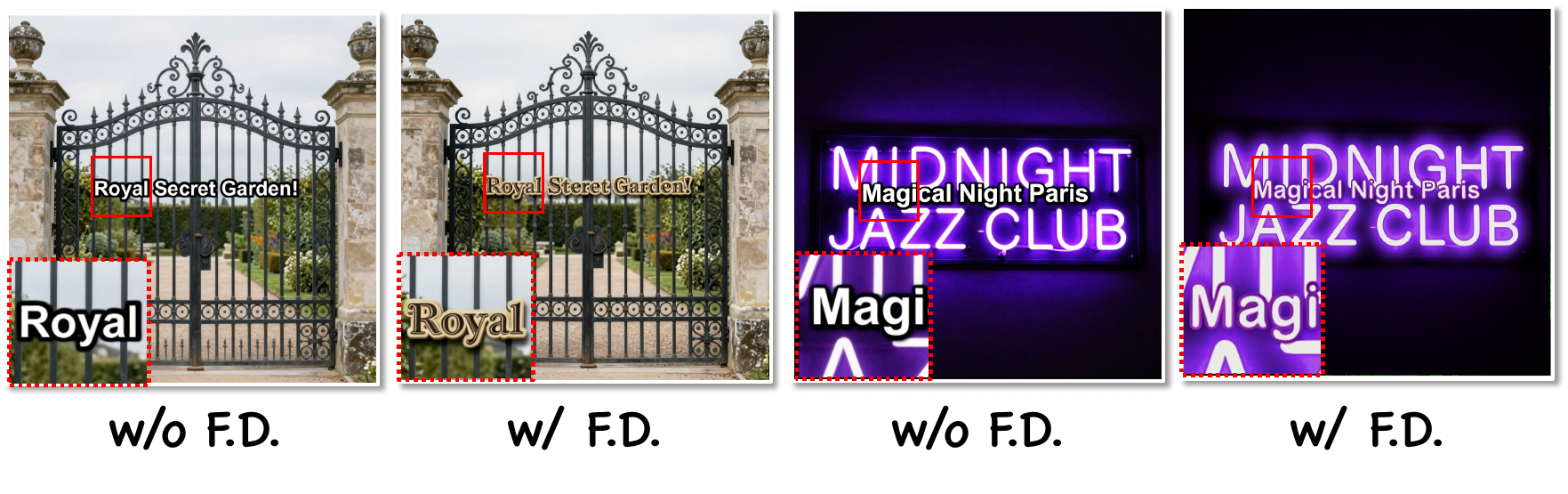}
    \caption{Qualitative comparisons for illustrating methods of Frequency Decomposition.}
    \label{fig:qualitative_fd}
\end{figure}

\noindent\textbf{Ablation study of F.D. in latent space.}
Fig.~\ref{fig:qualitative_fd} shows the impact of F.D. for improving the text rendering quality. During rendering, the unwanted dark edges persist alongside the text strokes, as can be seen in the contours of 'Royal' and 'Magi' without F.D.. With F.D., the text is rendered more harmoniously with the background than without F.D. 
It illustrates that F.D. preserves space for style and color while maintaining the text structure. In addition, Table~\ref{tab:quantitative_ablation} demonstrates metrics across precision faces comprehensive decline without F.D..

\noindent\textbf{Ablation study of Injection.}
It can be observed from Fig.~\ref{fig:qualitative_ablation} that the glyph injection significantly improves the text precision, which can be verified by OCR scores shown in Table~\ref{tab:quantitative_ablation}. It demonstrates that leveraging the glyph information to re-weight attention value, and injecting the glyph latent into the latent space gain significant improvement on text rendering precision.

\noindent\textbf{Ablation study of Iterative Refine.}
The iterative refine process is shown in the right side of Fig.~\ref{fig:qualitative_ablation}, which significantly improves the text rendering quality, and trends can be visual by the Fig.~\ref{fig:quantitative_refine}. This process contributes to improving style score without harming the rendering accuracy. It indicates that iterative refinement steadily enhances the Visual Quality of the rendered text while largely preserving Text Accuracy, demonstrating the effectiveness of our Style Refinement.

\section{Conclusion}
\label{sec:con}

We present GlyphBanana, a training-free agentic framework that bridges font-level precision and diffusion-model flexibility via frequency-decomposed latent injection, attention re-weighting, and VLM-driven iterative refinement.
Without any fine-tuning, it generalises across DiT backbones and surpasses all baselines in both rendering accuracy and visual quality.
We further contribute GlyphBanana-Bench, the first benchmark covering common words, rare characters, and complex scientific formulas.

{
    \small
    \bibliographystyle{ieeenat_fullname}
    \bibliography{main}
}

\clearpage
\appendix
\section{More Qualitative Results}
\label{app:qualitative-results}
\begin{figure*}[p]
    \centering
    \includegraphics[width=\linewidth]{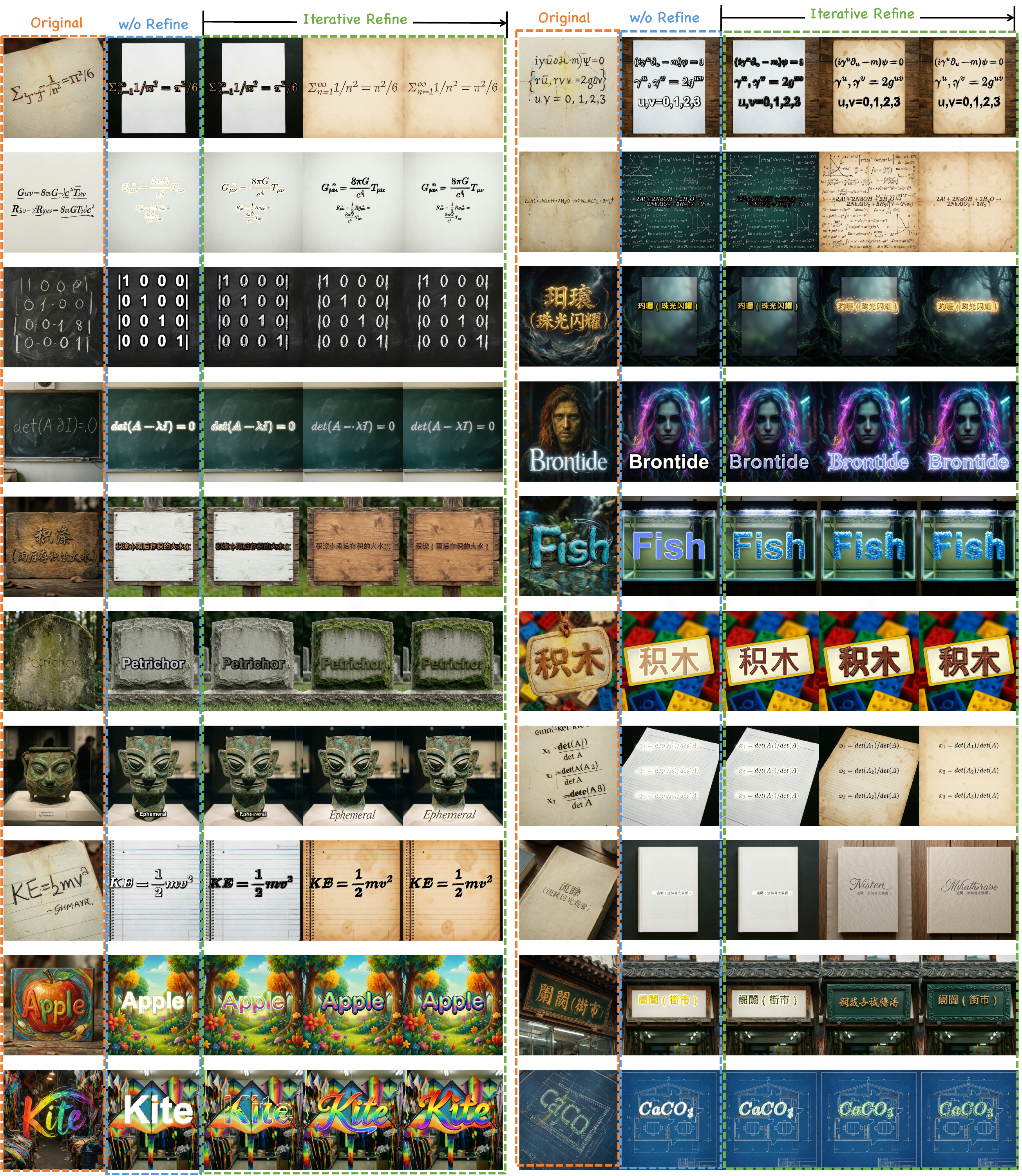}
    \caption{More qualitative results for refinement process.}
    \label{fig:1}
\end{figure*}

\begin{figure*}[p]
    \centering
    \includegraphics[width=\linewidth]{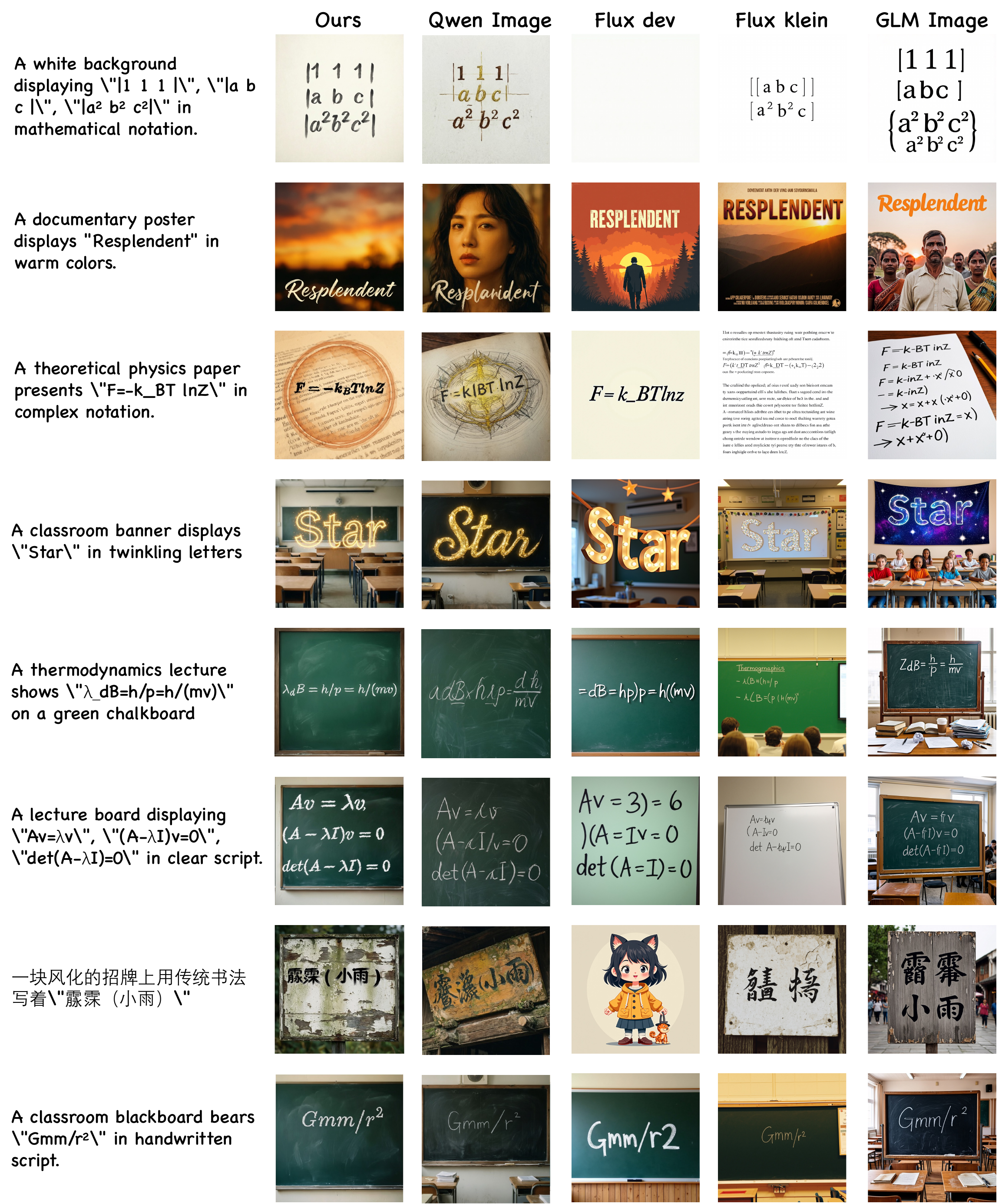}
    \caption{More qualitative results for GlyphBanana, Qwen-Image as base model.}
    \label{fig:2}
\end{figure*}

\begin{figure*}[p]
    \centering
    \includegraphics[width=\linewidth]{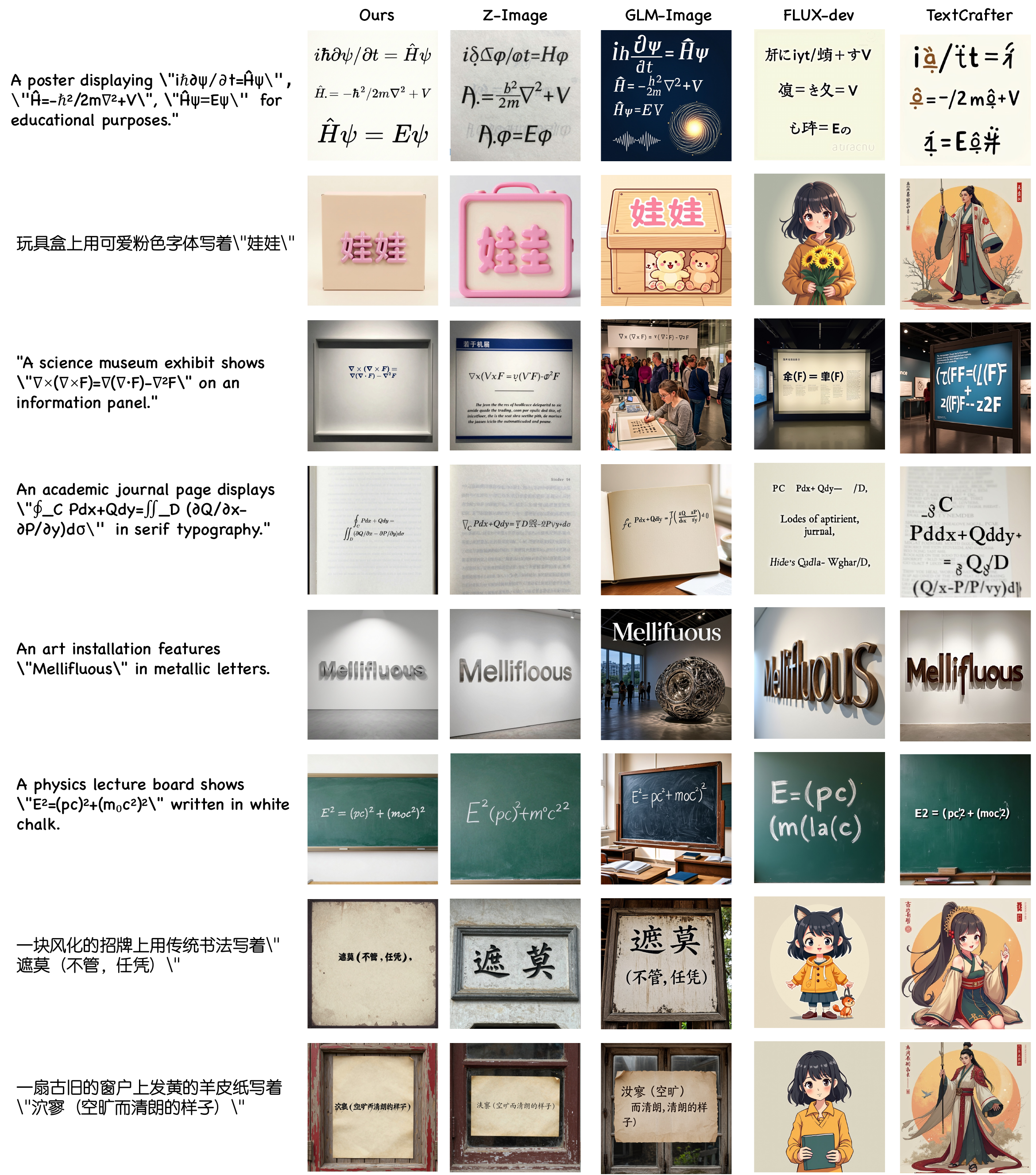}
    \caption{More qualitative results for GlyphBanana, Z-Image as base model.}
    \label{fig:3}
\end{figure*}

This section provides additional qualitative examples as shown in Fig.~\ref{fig:1},~\ref{fig:2},~\ref{fig:3} for visual comparison. 

\clearpage
\clearpage

\clearpage

\clearpage

\section{Benchmark Statistics}
\subsection{GlyphBanana-Benchmark Overview}
Table~\ref{tab:GlyphBanana_stats} reports the detailed statistics of GlyphBanana-Benchmark. The benchmark spans English, Chinese, and scientific-formula subsets, and the formula branch follows a ladder-shaped difficulty schedule from short expressions to long multi-line structures. This progression is useful for stress-testing both the layout planner and the auxiliary rendering tools under increasingly complex text lengths and prompt conditions.

\begin{table}[t]
\centering
\caption{\textbf{Illustration of GlyphBanana-Benchmark.}
The benchmark contains multimodal inputs spanning English, Chinese, and scientific-formula subsets, together with reference images and masks, and follows a ladder-shaped difficulty design. $Avg.|Text|$ denotes the average length of the target rendered text, and $Avg.|Prompt|$ denotes the average length of the corresponding prompt.}
\label{tab:GlyphBanana_stats}
\setlength{\tabcolsep}{5pt}
\resizebox{0.98\columnwidth}{!}{
\begin{tabular}{lccc}
\toprule
Subset & Num. & $Avg.|Text|$ & $Avg.|Prompt|$ \\
\midrule
\multicolumn{4}{l}{\textit{English Subsets}} \\
\quad GlyphBanana-En (Easy) & 50 & 4.08 & 47.74 \\
\quad GlyphBanana-En (Rare) & 25 & 8.92 & 56.84 \\
\midrule
\multicolumn{4}{l}{\textit{Chinese Subsets}} \\
\quad GlyphBanana-Zh (Easy) & 50 & 2.00 & 19.00 \\
\quad GlyphBanana-Zh (Rare) & 25 & 11.20 & 27.48 \\
\midrule
\multicolumn{4}{l}{\textit{Scientific Subsets (Ladder Difficulty)}} \\
\quad GlyphBanana-F (Easy) & 35 & 6.46 & 62.46 \\
\quad GlyphBanana-F (Mid) & 45 & 16.04 & 72.64 \\
\quad GlyphBanana-F (Hard L1) & 40 & 41.41 & 94.12 \\
\quad GlyphBanana-F (Hard L2) & 20 & 303.35 & 376.85 \\
\midrule
\textbf{Total / Average} & \textbf{290} & \textbf{32.68} & \textbf{76.62} \\
\bottomrule
\end{tabular}
}
\end{table}

\section{Layout Planner Agent}
\subsection{VLM-Based Text Grounding with Auxiliary Tools}
Table~\ref{tab:text_grounding} summarizes the ablation study for the layout planner, where IoU measures the overlap between predicted and ground-truth bounding boxes for formula placement. The \textbf{VLM only} setting uses the VLM without any coordinate-grid overlay as the baseline, while the remaining variants equip the same planner with grids of different densities. The key conclusion is that moderate-density coordinate aids are most effective: adding a 5$\times$5 grid raises mean IoU from 0.2703 to 0.5531, corresponding to a 104.6\% improvement over the VLM-only baseline. In contrast, the 8$\times$8 grid achieves only a 39.7\% gain, suggesting that overly dense visual guides introduce clutter and weaken spatial grounding. This observation motivates the current planner design, which uses a coordinate overlay to improve spatial grounding while explicitly instructing the VLM to ignore the red guide lines when describing scene content.
\begin{table}[t]
    \centering
    \caption{Ablation Study on VLM-Based Text Grounding with Auxiliary Tools.}
    \label{tab:text_grounding}
    \small
    \setlength{\tabcolsep}{1pt}
    \begin{tabular}{lcccc}
    \toprule
    \textbf{Configuration} & \textbf{Mean IoU$\uparrow$} & \textbf{Median IoU$\uparrow$} & \textbf{Std$\downarrow$} & \textbf{Improvement} \\
    \midrule
    VLM only) & 0.2703 & 0.2508 & 0.1620 & -- \\
    \midrule
    VLM + 3$\times$3 Grid & 0.4475 & 0.3892 & 0.1464 & +65.6\% \\
    \rowcolor{gray!15}
    \textbf{VLM + 5$\times$5 Grid} & \textbf{0.5531} & \textbf{0.5406} & \textbf{0.1280} & \textbf{+104.6\%} \\
    VLM + 8$\times$8 Grid & 0.3776 & 0.3628 & 0.1552 & +39.7\% \\
    \bottomrule
    \end{tabular}
\end{table}

\subsection{Formula Renderer as an Auxiliary Tool}
The formula renderer provides a deterministic auxiliary tool for synthesizing the glyph template used by the downstream injection stage. According to the implementation in \texttt{infer/formula\_helper.py}, the tool first detects whether the input should be treated as mathematical content, converts Unicode math symbols into LaTeX-compatible expressions when needed, performs lightweight automatic line breaking for long expressions, and then dispatches the content to a renderer selected by capability. For LaTeX-like content, the preferred route is MathJax through a Node.js backend; if that path is unavailable, the system falls back to \texttt{matplotlib} mathtext; otherwise, plain text is rendered with PIL and a font selected from the available registry. Multi-line expressions are rendered line by line and then vertically composed, after which the final glyph canvas can optionally be rotated to match the planned layout.

\begin{figure*}[t]
\centering
\resizebox{0.6\linewidth}{!}{
\begin{tikzpicture}[
    node distance=10mm and 18mm,
    every node/.style={font=\small},
    io/.style={rectangle, rounded corners, draw=black, fill=blue!6, align=center, text width=3.8cm, minimum height=1.0cm},
    proc/.style={rectangle, rounded corners, draw=black, fill=green!6, align=center, text width=4.0cm, minimum height=1.05cm},
    dec/.style={diamond, aspect=2.2, draw=black, fill=yellow!10, align=center, text width=2.6cm, inner sep=1pt},
    line/.style={-{Latex[length=2mm]}, thick}
]
\node[io] (input) {Input text + bbox\\(+ color / weight / font)};
\node[proc, below=of input] (convert) {Unicode-to-LaTeX conversion\\and auto line breaking};
\node[dec, below=12mm of convert] (latex) {LaTeX-like\\content?};
\node[dec, below left=14mm and 22mm of latex] (mathjax) {MathJax\\available?};
\node[proc, below=12mm of mathjax] (mj) {Render with MathJax\\SVG $\rightarrow$ PNG};
\node[proc, right=20mm of mj] (mpl) {Fallback to\\matplotlib mathtext};
\node[proc, right=24mm of latex] (pil) {Render plain text\\with PIL + font registry};
\node[proc, below=18mm of mpl] (compose) {Compose to target canvas,\\stack multi-line results, apply rotation};
\node[io, below=of compose] (output) {Glyph template image};

\draw[line] (input) -- (convert);
\draw[line] (convert) -- (latex);
\draw[line] (latex.east) -- node[above] {No} (pil.west);
\draw[line] (latex.west) -- node[above] {Yes} (mathjax.east);
\draw[line] (mathjax) -- node[left] {Yes} (mj);
\draw[line] (mathjax.east) -- ++(8mm,0) |- node[pos=0.25,right] {No} (mpl.north);
\draw[line] (mj.south) |- (compose.west);
\draw[line] (mpl.south) -- (compose);
\draw[line] (pil.south) |- (compose.east);
\draw[line] (compose) -- (output);
\end{tikzpicture}
}
\caption{Execution flow of the formula-rendering auxiliary tool used by GlyphBanana. The implementation prefers MathJax for rich LaTeX formulas, falls back to \texttt{matplotlib} mathtext when Node.js or SVG conversion is unavailable, and uses PIL-based text rendering for non-LaTeX content.}
\label{fig:formula_renderer_flow}
\end{figure*}
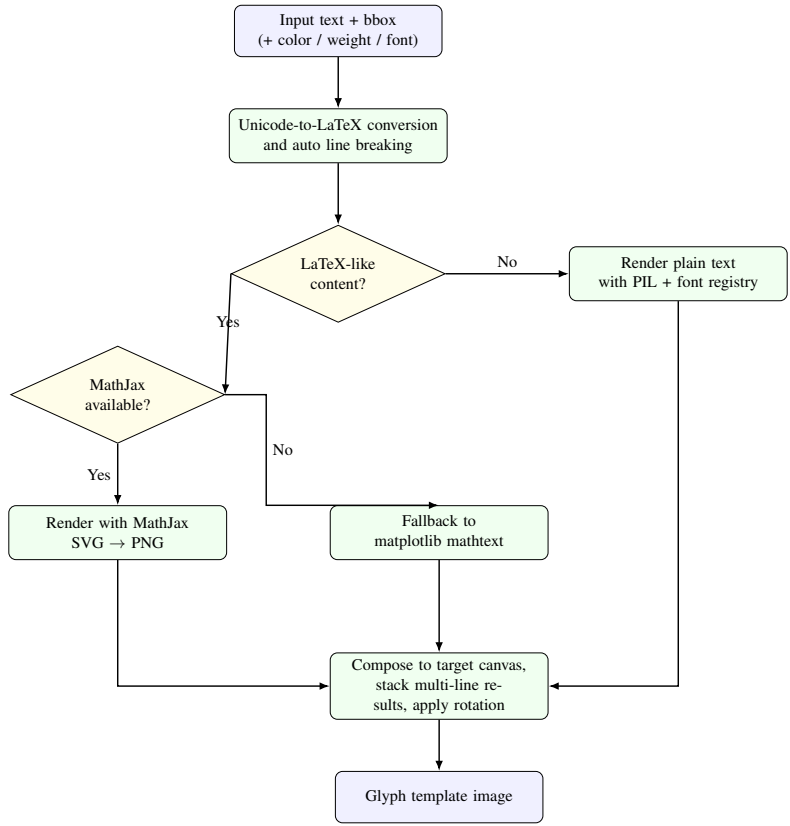

\subsection{Glyph Template Injection Illustration}
\begin{figure*}[t]
    \centering
    \includegraphics[width=0.8\linewidth]{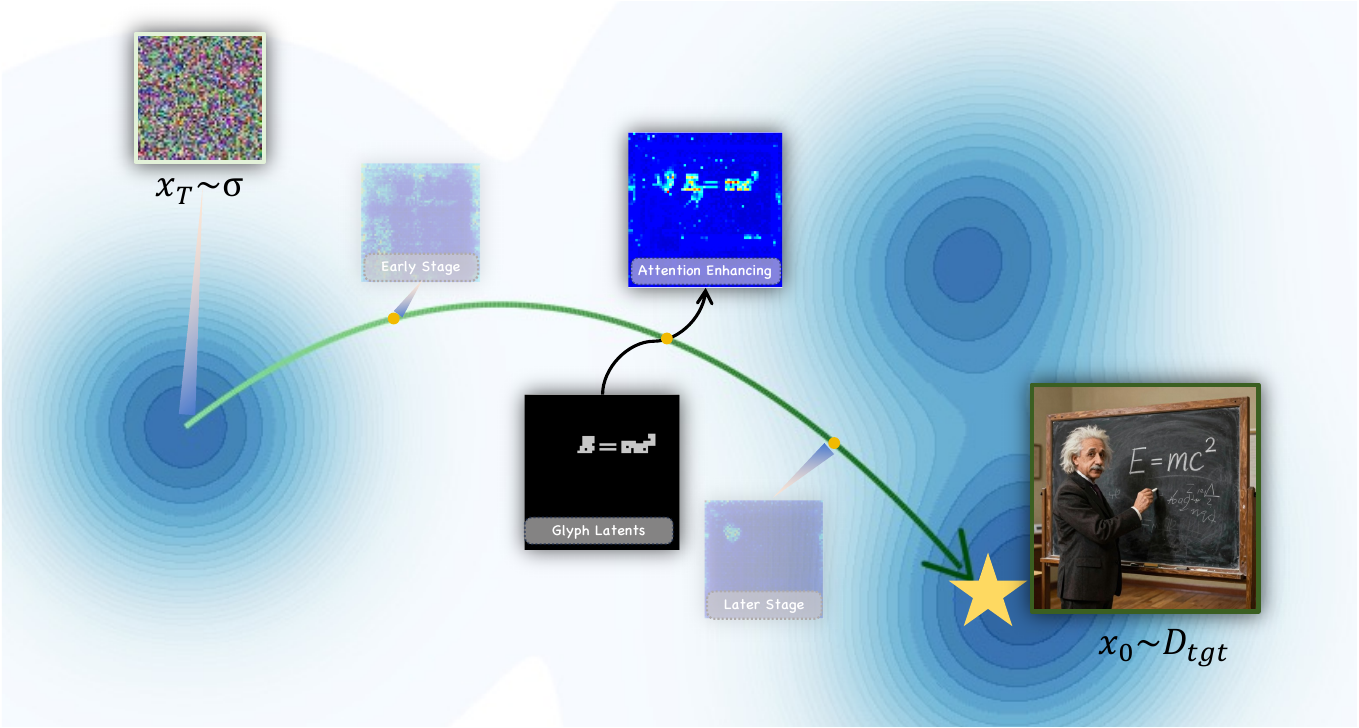}
    \caption{Schematic diagram of enhanced text rendering by injecting glyph templates in latent space.}
    \label{fig:glyphinject}
\end{figure*}

Once the auxiliary renderer produces a glyph template aligned with the typography plan, GlyphBanana injects that template into the latent-space refinement process. Figure~\ref{fig:glyphinject} visualizes this stage and complements the main-paper method description by showing how the rendered glyph prior is fused with the diffusion latent while preserving the surrounding scene structure.

\section{VLM Agent Prompt Templates}
\label{app:prompt-templates}
This section documents the prompt templates used by the current VLM agent implementation. The prompt stack is designed to be model-agnostic and can be attached to different diffusion backbones as long as they support the required conditioning interfaces. In a representative workflow, a text-to-image diffusion model first produces a reference image, after which the VLM planner infers a structured typography plan from the reference image, the user prompt, and the target text contents. The clean-prompt and style-prompt modules then support background regeneration, glyph injection, and subsequent harmonization with an image-to-image diffusion model.

\subsection{Typography Analysis Prompt}
\paragraph{Scenario.}
This prompt is invoked after Stage~2 reference-image generation and before any glyph injection is prepared. It is used only when the user does not manually override text regions.

\paragraph{Function.}
Its role is to transform an unstructured visual reference into a machine-readable typography plan that specifies both global scene attributes and per-region rendering instructions.

\paragraph{Inputs and outputs.}
The call consumes four pieces of information: the Stage~2 reference image, the original user prompt, the list of text or formula contents to be rendered, and a dynamically generated font list. The returned output is a strict JSON object with two top-level fields, \texttt{image\_analysis} and \texttt{text\_regions}. The former provides scene-level descriptors such as background style, dominant colors, and text style hints; the latter provides region-level attributes such as bounding boxes, font choice, color, alignment, and rotation.

\paragraph{Dependencies.}
The prompt depends on the grid-overlay utility, the font registry exposed by \texttt{infer/formula\_helper.py}, and the VLM backend configured in \texttt{VLMAgent}. The resulting typography plan is later consumed by both the glyph injector and the Stage~4 style harmonizer.

\begin{tcolorbox}[promptbox,title=Prompt: Typography Analysis]
\small
You are an expert in image typography analysis. Given a reference image with a 5$\times$5 grid and coordinate annotations, analyze the natural text rendering style and overall scene. Then plan the best typography layout for each text/formula item.

\vspace{0.5em}
\noindent\textbf{Critical constraints.}
\begin{itemize}[leftmargin=1.5em,nosep]
    \item CRITICAL: The reference image shows text that is FLAT and FACING the screen directly (frontal view, no perspective distortion). The planned boxes must also remain flat and frontal, with parallel top and bottom edges and no angled or perspective-distorted regions.
    \item IMPORTANT: The red grid lines and coordinate labels are ONLY positioning aids added by the system. They are not part of the actual image and must be ignored when describing \texttt{background\_style}, \texttt{dominant\_colors}, and \texttt{text\_style\_hint}.
    \item The 5$\times$5 grid uses the coordinates $\{0.0,0.2,0.4,0.6,0.8,1.0\}$ on each axis and provides the positioning reference for normalized coordinates in $[0,1]$.
\end{itemize}

\vspace{0.5em}
\noindent\textbf{Per-region fields.}
\begin{itemize}[leftmargin=1.5em,nosep]
    \item \texttt{content}: target text or formula.
    \item \texttt{bbox}: $[x_{\min},y_{\min},x_{\max},y_{\max}]$ in $[0,1]$, flat and horizontal.
    \item \texttt{font}: a font selected from the registered font list, or \texttt{auto}.
    \item \texttt{font\_weight}: \texttt{light}/\texttt{regular}/\texttt{bold}.
    \item \texttt{font\_size\_ratio}: scalar in $[0.1,1.0]$ relative to box height.
    \item \texttt{color}: one of \texttt{white}, \texttt{black}, \texttt{red}, \texttt{blue}, \texttt{green}, \texttt{yellow}, \texttt{orange}, \texttt{brown}, \texttt{gray}, \texttt{gold}, \texttt{silver}, \texttt{purple}, \texttt{pink}.
    \item \texttt{is\_latex}: boolean flag indicating whether the region is a formula.
    \item \texttt{alignment}: \texttt{left}/\texttt{center}/\texttt{right}.
    \item \texttt{rotation}: rotation angle in degrees, with $0$ denoting horizontal text.
\end{itemize}

\vspace{0.5em}
Available fonts are provided dynamically through the placeholder \texttt{\{font\_list\}}. The output is required to be strict JSON with two top-level entries, \texttt{image\_analysis} and \texttt{text\_regions}.
\end{tcolorbox}

\subsection{Generate Clean Prompt}
\paragraph{Scenario.}
This prompt is called at the beginning of Stage~3, after the typography plan has already been produced and immediately before background denoising and glyph injection.

\paragraph{Function.}
Its role is to remove explicit text-rendering instructions from the original prompt so that the diffusion backbone can focus on regenerating a clean background rather than hallucinating additional text.

\paragraph{Inputs and outputs.}
The interface accepts the original prompt and optionally a typography plan. In the current \texttt{VLMAgent} implementation, the function signature still exposes \texttt{typography\_plan}, but the active call path only forwards the original prompt text into the VLM prompt body. The output is a single rewritten clean prompt string.

\paragraph{Dependencies.}
This prompt depends on the original user prompt and the VLM backend. Its output is then fed directly into the Stage~3 denoising step, where it conditions the background generation used for subsequent pixel-space text compositing and latent injection.

\begin{tcolorbox}[promptbox,title=Prompt: Generate Clean Prompt]
\small
Remove ALL quoted text, formulas, and text-rendering instructions from the prompt. Keep ONLY the scene/background/style description. Add ``no text visible'' at the end.

\vspace{0.5em}
\noindent\textbf{Examples.}
\begin{itemize}[leftmargin=1.5em,nosep]
    \item Input: A classroom blackboard displays ``E=mc²'' in elegant chalk writing.\\
    Output: An empty classroom blackboard as background, clear and without any text. No text visible.
    \item Input: A stone monument is engraved with ``Knowledge is power'', surrounded by a bamboo grove.\\
    Output: A blank stone monument as background, surrounded by a bamboo grove, clear and without any text. No text visible.
\end{itemize}

Output ONLY the cleaned prompt, nothing else.
\end{tcolorbox}

\subsection{Generate Style Prompt}
\paragraph{Scenario.}
This prompt is used in Stage~4 when an image-to-image diffusion model is employed for style harmonization. It is called after Stage~3 has produced the injected image and after the planner has already produced \texttt{image\_analysis}.

\paragraph{Function.}
Its purpose is to compress the scene-level analysis into a short editing instruction that preserves the background while restyling the foreground text or formulas so that they better harmonize with the image.

\paragraph{Inputs and outputs.}
The input is the \texttt{image\_analysis} field of the typography plan, specifically the background style, dominant colors, and text-style hint. The output is a short English editing prompt, typically 10--30 words, which is then forwarded to the image-to-image diffusion model used for harmonization.

\paragraph{Dependencies.}
This prompt depends on the success of the typography-analysis stage, because it reuses the planner's scene descriptors instead of reading the image again. Its downstream dependency is the Stage~4 image-to-image diffusion model, which consumes the resulting editing instruction as its conditioning prompt.

\begin{tcolorbox}[promptbox,title=Prompt: Generate Style Prompt]
\small
You generate a SHORT image-editing instruction (10--30 words) for a style-transfer model. Goal: restyle foreground text to harmonize with the background while keeping the background untouched. Do NOT move, resize, or alter any text content or position.

\vspace{0.5em}
\noindent\textbf{Examples.}
\begin{itemize}[leftmargin=1.5em,nosep]
    \item Input:\\ \texttt{background\_style="weathered stone wall"},\\ \texttt{colors=["\#8B7D6B","\#A09080"]},\\ \texttt{hint="carved stone lettering"}\\
    Output: Restyle text as deeply carved stone engravings matching the weathered wall texture and earthy tones.
    \item Input:\\ \texttt{background\_style="neon-lit cyberpunk street"},\\ \texttt{colors=["\#FF00FF","\#00FFFF"]}, \texttt{hint="glowing neon sign"}\\
    Output: Make text glow like neon signs with magenta and cyan edges against the dark street scene.
    \item Input:\\ \texttt{background\_style="minimalist white paper"},\\ \texttt{colors=["\#FFFFFF","\#E0E0E0"]},\\ \texttt{hint="clean printed type"}\\
    Output: Render text as crisp black ink print on the clean white background with subtle shadow.
\end{itemize}

Output ONLY the instruction, nothing else. It must be in English and 10--30 words long.
\end{tcolorbox}

\subsection{Refine Prompt}
\paragraph{Scenario.}
This prompt is a reserved interface in \texttt{VLMAgent} for generic prompt enhancement. It is optional in the overall pipeline and may be enabled or disabled depending on the target diffusion backbone and deployment strategy.

\paragraph{Function.}
Its goal is to rewrite a user prompt into a more rendering-friendly form while preserving the quoted text exactly.

\paragraph{Inputs and outputs.}
The interface accepts the original prompt, an optional text-content hint, the number of variants to sample, and a temperature value. It returns one or more rewritten prompt strings.

\paragraph{Dependencies.}
The prompt depends only on the VLM backend. In some deployments, this refinement step may be replaced by deterministic prompt normalization, so the VLM-based refiner remains optional rather than mandatory.

\begin{tcolorbox}[promptbox,title=Prompt: Refine Prompt]
\small
You are a prompt engineer for a text-to-image model that renders text inside images.
\begin{enumerate}[leftmargin=1.5em,nosep]
    \item Keep the original scene description and ALL quoted text exactly as-is.
    \item Add that text should be clearly legible, well-positioned, and high-contrast.
    \item Add brief visual details (lighting, style, materials) that make the scene vivid.
\end{enumerate}
Output only the enhanced prompt, nothing else.
\end{tcolorbox}

\subsection{Score Image Prompt}
\paragraph{Scenario.}
This prompt defines a generic absolute image scorer in \texttt{VLMAgent}. It is not required by the core pipeline, because many deployments instead rely on OCR-based selection or external evaluation metrics.

\paragraph{Function.}
Its purpose is to assign a single scalar score to one generated image by jointly considering image quality, prompt alignment, and text readability.

\paragraph{Inputs and outputs.}
The interface takes one image together with the corresponding prompt and an optional explicit text-content string. It returns a single floating-point score in the range $[0,10]$.

\paragraph{Dependencies.}
This prompt depends on the VLM backend and a parsed image input. It is kept as a reusable evaluation primitive for alternative pipelines, future ablations, or backbone-specific selection strategies.

\begin{tcolorbox}[promptbox,title=Prompt: Score Image]
\small
Rate this image 0--10 based on:
\begin{itemize}[leftmargin=1.5em,nosep]
    \item overall quality (clarity, color, composition): 0--3;
    \item alignment with prompt: 0--4;
    \item text accuracy and readability (if applicable): 0--3.
\end{itemize}
Output only the numeric score, nothing else.
\end{tcolorbox}

\subsection{Rank Images Prompt}
\paragraph{Scenario.}
This prompt defines a generic multi-image ranking interface in \texttt{VLMAgent}. Similar to the single-image scorer, it is optional and can be switched on when a deployment prefers VLM-based ranking over OCR-based candidate selection.

\paragraph{Function.}
Its role is to sort several candidate images from best to worst under shared criteria, so that rank positions can be converted into stepwise scores.

\paragraph{Inputs and outputs.}
The call accepts a list of candidate images, the original prompt, and optionally the expected text string. It returns an ordered index list, which the implementation then maps to descending scores.

\paragraph{Dependencies.}
The prompt depends on a multi-image VLM call. In many text-rendering settings, this functionality is superseded by OCR-based selection, which is more directly tied to rendering precision.

\begin{tcolorbox}[promptbox,title=Prompt: Rank Images]
\small
Rank these \{n\} images from best to worst based on overall quality, prompt alignment, and text accuracy (if applicable). Output only the ranking as comma-separated indices (for example, \texttt{3,1,4,2}), nothing else.
\end{tcolorbox}

\section{Evaluation Interfaces}
This section summarizes the prompt-based evaluation interfaces associated with the VLM agent. In a representative model-agnostic deployment, candidate selection can be performed by OCR-based scoring, which asks the VLM to transcribe the rendered text and compares the result against the target string extracted from the prompt. Other interfaces, including style scoring, faithfulness scoring, VQAScore, and CLIPScore, remain reusable evaluation components that can be enabled or disabled depending on the evaluation protocol.

\subsection{API-Based OCR Recognition}
\paragraph{Scenario.}
This prompt is used for final candidate selection after the reference image, injected result, and harmonized variants have all been generated.

\paragraph{Function.}
Its role is to directly transcribe the visible text from each candidate image so that the system can compare recognized text against the target content and choose the most accurate rendering.

\paragraph{Inputs and outputs.}
For the API-based VLM evaluator, the expected text $T$ is extracted from the quoted spans in the input prompt. Each call then consumes one candidate image and returns a raw recognized text string with no explanation. The resulting transcription is subsequently compared with $T$ using edit-distance metrics.

\paragraph{Dependencies.}
This prompt depends on the image candidates produced by the generation pipeline, the quote-based text extractor in \texttt{eval/core/metrics.py}, and the VLM backend. It is a natural selector for model-agnostic text-rendering systems because it directly measures rendered-text fidelity rather than relying on backbone-specific confidence signals.

\noindent The model is asked to directly transcribe the rendered text:

\begin{tcolorbox}[promptbox,title=Prompt: OCR Recognition]
\small
Please read and output ALL the text content visible in this image. Only output the text you can see, nothing else. If there are multiple text elements, separate them with spaces. Do not add any explanations or descriptions, just the raw text content.
\end{tcolorbox}

Let $\mathcal{N}(\cdot)$ denote lowercase normalization with whitespace collapsing, and let $d_{\mathrm{Lev}}$ be the Levenshtein distance. If $R$ is the recognized text, the VLM-based text scores are computed as
\begin{equation}
\mathrm{Acc}_{\mathrm{VLM}}=\max\!\left(0,1-\frac{d_{\mathrm{Lev}}(\mathcal{N}(T),\mathcal{N}(R))}{|\mathcal{N}(T)|}\right),
\end{equation}
\begin{equation}
\mathrm{NED}_{\mathrm{VLM}}=\max\!\left(0,1-\frac{d_{\mathrm{Lev}}(\mathcal{N}(T),\mathcal{N}(R))}{\max\left(|\mathcal{N}(T)|,|\mathcal{N}(R)|\right)+\varepsilon}\right),
\end{equation}
where $\varepsilon$ is a small constant for numerical stability.

For the standalone OCR metric used in the benchmark tables, we additionally report MinerU-based OCR scores with the same edit-distance formulation after normalizing the recognized text.

\subsection{VLM Style Score}
\paragraph{Scenario.}
This prompt belongs to the evaluation toolkit rather than the active generation path.

\paragraph{Function.}
It estimates image-level style and quality compatibility through a direct scalar judgment on a 0--10 scale.

\paragraph{Inputs and outputs.}
The interface takes a single generated image as input and returns one scalar score, which is normalized into $S_{\mathrm{style}}$.

\paragraph{Dependencies.}
It depends only on the VLM evaluation backend and does not require the original prompt text.

\noindent The VLM style score is implemented as a direct quality judgment on a 0--10 scale:

\begin{tcolorbox}[promptbox,title=Prompt: VLM Style Score]
\small
Evaluate the overall quality of this image considering:
\begin{enumerate}[leftmargin=1.5em,nosep]
    \item image clarity and sharpness;
    \item visual coherence and aesthetics;
    \item proper rendering of all elements.
\end{enumerate}
Rate from 0--10, respond with only a number.
\end{tcolorbox}

If the returned scalar is $s_{\mathrm{style}}\in[0,10]$, we normalize it as $S_{\mathrm{style}}=\frac{s_{\mathrm{style}}}{10}$.

\subsection{VLM Faithfulness Score}
\paragraph{Scenario.}
This prompt is used in the evaluation module to quantify prompt adherence, but it is not invoked during the main generation loop.

\paragraph{Function.}
Its goal is to measure whether the generated image remains faithful to the full prompt, including scene description, object presence, style intent, and text placement.

\paragraph{Inputs and outputs.}
The interface consumes one generated image together with the original prompt and returns one scalar score in the range $[0,10]$, which is then normalized into $S_{\mathrm{faith}}$.

\paragraph{Dependencies.}
It depends on both the image and the original prompt text, because faithfulness is defined relative to the complete semantic condition rather than OCR accuracy alone.

\noindent Prompt faithfulness is measured by asking the VLM to jointly assess scene consistency, object completeness, style fidelity, and text placement:

\begin{tcolorbox}[promptbox,title=Prompt: Faithfulness Score]
\small
You are evaluating how faithfully this generated image matches its text prompt.

Prompt: ``\textit{original prompt}''

Consider the following aspects:
\begin{enumerate}[leftmargin=1.5em,nosep]
    \item Scene \& background: does the scene match the description?
    \item Objects \& elements: are all described objects and elements present?
    \item Style \& color: does the visual style match the prompt's intent?
    \item Text content \& placement: is the text rendered in the correct location with correct content?
\end{enumerate}
Rate the overall faithfulness from 0--10, respond with only a number.
\end{tcolorbox}

If the raw response is $s_{\mathrm{faith}}\in[0,10]$, we use the normalized score $S_{\mathrm{faith}}=\frac{s_{\mathrm{faith}}}{10}$.

\subsection{VQAScore Interface}
\paragraph{Scenario.}
This interface is part of the evaluation stack and is independent of the generation-time prompt calls in the model-agnostic pipeline.

\paragraph{Function.}
Its role is to compute a paired image-text relevance score without additional prompt engineering.

\paragraph{Inputs and outputs.}
The interface takes the generated image path and the original prompt $P$ as its text query. The output is a scalar relevance score returned by the \texttt{clip-flant5-xxl}-based VQAScore model.

\paragraph{Dependencies.}
It depends on the local VQAScore wrapper under \texttt{eval/TextCrafter\_Eval/vqascore.py} and the underlying \texttt{t2v\_metrics} implementation.

\noindent For VQAScore, we do not perform extra prompt engineering. Instead, the original prompt $P$ is directly used as the text query paired with image $I$ in the \texttt{clip-flant5-xxl} scorer:

\begin{tcolorbox}[promptbox,title=Text Query for VQAScore]
\small
Input text to VQAScore: the original prompt $P$ itself, without additional instructions or template wrapping.
\end{tcolorbox}

The resulting score is
\begin{equation}
S_{\mathrm{VQA}} = f_{\mathrm{VQA}}(I,P),
\end{equation}
where $f_{\mathrm{VQA}}$ denotes the paired image-text score returned by the VQAScore model.

\subsection{CLIP Score}
\paragraph{Scenario.}
This metric is used only in the evaluation module and is not part of the runtime prompt workflow of the model-agnostic generation pipeline.

\paragraph{Function.}
It measures global image-text alignment between the generated result and the original prompt.

\paragraph{Inputs and outputs.}
The implementation takes one image path and one prompt string, prepends the fixed text prefix ``A photo depicts'' to the prompt, and returns a non-negative scalar CLIPScore.

\paragraph{Dependencies.}
It depends on the CLIP ViT-L/14 encoder loaded in \texttt{eval/core/metrics.py}. The image and text embeddings are normalized before their cosine similarity is rescaled into the final score.

\noindent For image $I$ and prompt $P$, CLIP produces image and text embeddings, denoted by $\phi_{\mathrm{img}}(I)$ and $\phi_{\mathrm{text}}(P)$. The implementation converts cosine similarity into a non-negative score via
\begin{equation}
S_{\mathrm{CLIP}} = 2.5 \cdot \max \!\left(
\frac{\phi_{\mathrm{img}}(I)^\top \phi_{\mathrm{text}}(P)}
{\|\phi_{\mathrm{img}}(I)\|_2\;\|\phi_{\mathrm{text}}(P)\|_2},\, 0
\right).
\end{equation}

\end{document}